\documentclass{article}

\usepackage{arxiv}

\usepackage[utf8]{inputenc} 
\usepackage[T1]{fontenc}    
\usepackage{hyperref}       
\usepackage{url}            
\usepackage{booktabs}       
\usepackage{nicefrac}       
\usepackage{microtype}      
\usepackage{lipsum}		
\usepackage{graphicx}
\usepackage{natbib}
\usepackage{doi}
\usepackage{amsmath,amssymb,amsfonts}%
\usepackage{amsthm}%
\usepackage{mathrsfs}%

\usepackage{subfig}

\usepackage{caption} 
\usepackage{multirow}

\usepackage{algorithm}
\usepackage{algpseudocode}
\usepackage{mathtools}

\title{Boundary Attention Mapping (BAM): Fine-grained saliency maps for segmentation of Burn Injuries}

\author{ {Mahla ~Abdolahnejad} \textsuperscript{1,3,*}\\
    \texttt{mahla@skinopathy.com}\\
	\And
	{Justin ~Lee} \textsuperscript{1,2,*} \\
    \texttt{justinlee@ualberta.ca}\\
    \And
    {Hannah ~Chan} \textsuperscript{1,4,**}\\
    \texttt{hannah@skinopathy.com}\\
    \And
    {Alex ~Morzycki} \textsuperscript{1,2}\\
    \texttt{morzycki@ualberta.ca}\\
    \And
    {Olivier ~Ethier} \textsuperscript{5,6}\\
    \texttt{olivier.ethier@umontreal.ca}\\
    \And
    {Anthea ~Mo} \textsuperscript{1}\\
    \texttt{anthea@skinopathy.com}\\
    \And
    {Peter X.~Liu} \textsuperscript{3}\\
    \texttt{xpliu@sce.carleton.ca}\\
    \And
    {Joshua N.~Wong} \textsuperscript{1,2,**}\\
    \texttt{jnw@ualberta.ca}\\
    \And
    {Colin ~Hong} \textsuperscript{1,4,**}\\
    \texttt{collin@skinopathy.com}\\
    \And
    {Rakesh ~Joshi} \textsuperscript{1}\\
    \texttt{rakesh@skinopathy.com}
	\And
    \\
	\textsuperscript{1} Skinopathy Research, Skinopathy Inc., North York, Ontario, Canada.\\
    \textsuperscript{2} University of Alberta Hospital, Edmonton, Alberta, Canada.\\
    \textsuperscript{3} Department of Systems and Computer Engineering, Carleton University, Ottawa, Ontario, Canada.\\
    \textsuperscript{4} Scarborough Health Network, Centenary Hospital, Scarborough, Ontario, Canada.\\
    \textsuperscript{5} University of Montreal, Montreal, Quebec, Canada.\\
    \textsuperscript{6} Quebec AI Institute (MILA), Montreal, Quebec, Canada.\\
    \\
    \textsuperscript{*} Contributed Equally.\\
    \textsuperscript{**} Senior clinical authors.\\
}

\date{}

\renewcommand{\headeright}{}
\renewcommand{\undertitle}{}
\renewcommand{\shorttitle}{}

\begin{document}
\maketitle

\begin{abstract}
Burn injuries can result from mechanisms such as thermal, chemical, and electrical insults. A prompt and accurate assessment of burns is essential for deciding definitive clinical treatments. Currently, the primary approach for burn assessments, via visual and tactile observations, is approximately 60\%-80\% accurate. The gold standard is biopsy and a close second would be non-invasive methods like Laser Doppler Imaging (LDI) assessments, which have up to 97\% accuracy in predicting burn severity and the required healing time. In this paper, we introduce a machine learning pipeline for assessing burn severities and segmenting the regions of skin that are affected by burn. Segmenting 2D colour images of burns allows for the injured versus non-injured skin to be delineated, clearly marking the extent and boundaries of the localized burn/region-of-interest, even during remote monitoring of a burn patient. \\
We trained a convolutional neural network (CNN) to classify four severities of burns: SPF (superficial), SPT (superficial partial thickness), DPT (deep partial thickness), and FT (full thickness). We built a saliency mapping method, Boundary Attention Mapping (BAM), that utilises this trained CNN for the purpose of accurately localizing and segmenting the burn regions from skin burn images. We demonstrated the effectiveness of our proposed pipeline through extensive experiments and evaluations using two datasets; 1) A larger skin burn image dataset consisting of 1684 skin burn images of four burn severities, 2) An LDI dataset that consists of a total of 184 skin burn images with their associated LDI scans. The CNN trained using the first dataset achieved an average F1-Score of 78\% and micro/macro- average ROC of 85\% in classifying the four burn severities. Moreover, a comparison between the BAM results and LDI results for measuring injury boundary showed that the segmentations generated by our method achieved 91.60\% accuracy, 78.17\% sensitivity, and 93.37\% specificity.
\end{abstract}

\section{Introduction}
\label{sec1}

The skin is the largest human organ in the body with many associated critical functions. As such, burn injuries can damage the skin and lead to loss of these functions such as immunoprotection, thermoregulation, and maintenance of euvolemia. Burns can be caused by many mechanisms such as thermal, chemical, and electrical insults \cite{pencle2017first}. Superficial or first degree burns affect the epidermis only and usually heal with minimal clinical intervention within a 1-week time span. Second degree burns, which include superficial partial thickness (SPT) and deep partial thickness (DPT) burns, are deeper burns that may require clinical intervention. SPT extends to the upper layer of the dermis and DPT to the deeper dermal layers. Third degree burns, also termed full thickness (FT) burns are self-explanatory, in that the entire skin thickness, including epidermis and dermis, are burned and necrotic \cite{schaefer2021thermal}. Another essential metric for assessing a burn severity is the percentage of total body surface area (TBSA\%) that is affected by the burn. Accurate and prompt assessment of these two metrics is required to determine definitive clinical treatment, including resuscitative calculations and for surgical planning \cite{cirillo2019time}. 

Definitive burn severity assessments can be invasive, using biopsies to determine the depth of injury, or non-invasive, using imaging methods like Laser Doppler imaging (LDI) which are usually only available in larger burn trauma centres \cite{hop2013cost}. LDIs are considered the non-invasive gold standard in burn assessment, as the laser scans the skin and provides information on compromised blood flow. LDI can be over 97\% accurate when paired with expert clinical interpretation in determining the healing time of second degree burns \cite{pape2001audit}.

In addition to the accuracy of the burn assessment, the time it takes for the assessment to be available also plays an essential role in a patient’s recovery. If a rapid accurate assessment is available, better treatment decisions can be made resulting in faster recovery, reduced expenses, and also decreased risk of hospital acquired complications \cite{abubakar2020assessment}. Nonetheless, a burn specialist or an experienced clinician is often not available at the point of initial burn assessment. This shortage of experienced clinicians is more pronounced in remote areas or even in the community setting in urban centres \cite{khan2020burnt}. Given all these reasons, it is necessary to move towards an alternative way of assessing burns. Such an alternative method for burn evaluation needs to be rapid, easily accessible, low-cost, and consistent in terms of accuracy. 

Machine learning (ML) models, working in tandem with computer vision components, can provide an alternative, not only for initial burn assessment (including the severity and area affected) but also to track the healing of a burn injury \cite{ethier2022using}. Convolutional neural network (CNN) models can be trained on clinically annotated burn images, taken from a mobile device or SLR, to classify images by the severity of the burn \cite{abubakar2020assessment}. Additionally, information from various neural layers of the CNN in the form of saliency maps can be used to map the edges of localized burn injuries. Comparing the severity and region of the burn as identified by ML models to LDI data of those patients can provide evidence that validates such an ML system for clinical use.

In this study, we propose a CNN-based attention mapping system for localization and segmentation of the burned regions from skin burn images. These segmentations can be used to obtain an accurate and automatic burn boundary of a localized burn to determine TBSA\%. This study builds on the literature on class-discriminative visualisations for deep CNNs trained for classification or recognition tasks. Class-discriminative visualisation methods focus on locating specific features in an image that support a specific class label while excluding the features irrelevant to that specific label. One state-of-the-art method uses the Gradient-weighted Class Activation Mapping (Grad-CAM) \cite{selvaraju2017grad}. Grad-CAM builds on the fact that the neurons in the last convolutional layers of a CNN possess information on both high-level semantics and detailed spatial information since they scan for class-related information in the image to make a prediction. Grad-CAM uses the gradient information of a target class flowing into the last convolutional layer of the CNN in order to understand the importance of each neuron for a class prediction. As a result, Grad-CAM is able to produce a coarse-grained localization map highlighting the important regions in the image for predicting that class. Although these heatmaps are highly class-discriminative and localized, they do not produce fine-grained details that can have clinical relevance. 

We propose the Boundary Attention Mapping (BAM) method which uses the Grad-CAM heatmaps as an intermediate representation for the purpose of generating fine-grained burn localizations and segmentations. More specifically, given a dataset of 2D-color skin burn images, we first train a deep CNN model with this dataset to predict four burn severities. Once the classifier is trained, a coarse-grained localization of the burn area is obtained using Grad-CAM. BAM then utilizes the coarse-grained Grad-CAM visualizations along with the activations of the first convolutional layer of the deep CNN in order to create a high-resolution visualization that highlights the burn area. This visualization can in turn be used for creating a fine-grained segmentation of the burn area. 

To validate the clinical relevance of this system, we also created a binary image dataset from LDI scans and 2D images and compared the predictions of the CNN-BAM system with this benchmark dataset.

\section{Methodology}
\label{sec2}

\subsection{CNN Architecture and Training}
\label{sec2-1}

We utilized the pre-trained EfficientNet-B7 architecture \cite{tan2019efficientnet}, a Convolutional Neural Network (CNN) model as the base architecture for implementing the Boundary Attention Mapper (BAM) model. The EfficientNet-B7 architecture is trained using the ImageNet dataset \cite{deng2009imagenet} for performing object recognition. The top layer was removed, and multiple fully connected layers, followed by a 4-class SoftMax layer, were added to the pre-trained architecture. Additionally, drop-out regularizations (with a value of 0.3), are included in the fully connected layers to prevent over-fitting on the training dataset. 

We fine-tuned this architecture using the skin burn image dataset and the 5-fold cross-validation method. The CNN is trained for classifying burn degrees using the categorical cross-entropy loss function. The optimizer used for the training is an ADAM optimizer with an initial learning rate of 0.001. The learning rate was set to dynamically decrease by a factor of 0.5, if validation accuracy did not improve in a set number of training epochs. We observed that approximately at the 40th-epoch cycle, the validation accuracy reached a plateau. We used this trained EfficientNet-based classifier for predicting the burn severity and as the model for implementing the Boundary Attention Mapper (BAM) method. More specifically, the Grad-CAM heatmaps and first convolutional layer activations required for BAM's implementation are obtained using this trained classifier in the inference mode.

\subsection{Grad-CAM}
\label{sec2-2}

Given a deep CNN trained for an image classification or recognition task, the Grad-CAM method can be used to generate a visualization of the image regions that contribute the most to the deep CNN's decision \cite{selvaraju2017grad}. In other words, it provides a coarse-grained heatmap of the attention pixels used to make that decision. Briefly:

Assuming that the deep CNN has predicted the class c for an input image, Grad-CAM first computes the gradient of the score for the predicted class with respect to the activations of the kth convolutional layer $\frac{\partial y^c}{\partial A^k}$. It then performs average pooling of these gradients over neurons of the activations $A^k$ to obtain the neurons' importance weights as follows:

\begin{equation}
\alpha^c_k = \frac{1}{z} \sum_i \sum_j \frac{\partial y^c}{\partial A^k_{ij}}.
\end{equation}

More precisely, $\alpha^c_k$ denotes the importance of the activations $A^k$ for a target class $c$. Finally, Grad-CAM computes its heatmap by passing the weighted activations through a $ReLU$ function,

\begin{equation}
L^c_{GradCAM} = ReLU(\sum_k \alpha^c_k A^k).
\end{equation}

The ReLU function helps to keep only positive influences on the prediction by zeroing out the negative gradients. $L^c_{GradCAM}$ is therefore a coarse-grained heatmap, the size of which is equal to the size of channels in $A^k$. The Grad-CAM heatmap is often computed for the last convolutional layer, which has a small channel size since the class-related information is best captured by the last layer. Consequently, as mentioned previously, Grad-CAM is successful in achieving highly localized and class-discriminative visualizations, but the visualizations suffer from low resolution which makes it inappropriate for clinical-decision support, by itself.

\subsection{Boundary Attention Mapper (BAM)}
\label{sec2-3}

Here, we introduce the BAM system, a method for obtaining fine-grained heatmaps from the coarse-grained Grad-CAM. These fine-grained BAM heatmaps can be used to obtain high-resolution segmentations of burn areas from 2D colour images from patients. BAM's primary concept is based on the observation that activations of early layers of a deep CNN produce heatmaps with higher resolutions. In addition, heatmaps can be identified in these channels that highlight the same regions as the Grad-CAM heatmaps. BAM measures the correlations between the Grad-CAM heatmap and the activation channels of the first convolutional layer. It is therefore possible to find a heatmap of attention pixels that is of much higher resolution than the Grad-CAM heatmap by itself. BAM, therefore, proposes an approach for combining these heatmaps of the first layer activation channels, based on their similarity to the Grad-CAM heatmap, with the purpose of achieving a fine-grained visualization of burn regions. The details of this procedure are as follows:

\subsubsection{Generating high-resolution visualizations}
\label{sec2-3-1}

The primary goal of BAM is to find a high-resolution heatmap for an input image of a burn injury, which highlights the same image regions as the Grad-CAM heatmap. As a result, the burn regions stand out in such heatmaps and therefore are easily distinguishable from other image regions. To achieve this goal, BAM uses the correlation score as a measure of similarity between the high-resolution visualization and the low-resolution Grad-CAM heatmap. More specifically, it uses a greedy algorithm that iterates through every channel of the first convolutional layer activations multiple times. In each iteration it selects a channel, which when added to the average of previously selected channels, results in the maximal increase in correlation with the Grad-CAM heatmap. This operation is illustrated by the following equation:

\begin{equation}
ch_{idx} = \underset{ch}{\mathrm{argmax}}[\rho(A^1_{ch}, L^c_{GradCAM})],
\end{equation}

where $ch_{idx}$ is the selected channel, $ch$ is a list of combined channels, $A^1_{ch}$ is the heatmap computed by averaging the first layer activation channels in $ch$, and $\rho$ is Spearman's rank correlation coefficient between the Grad-CAM heatmap $L^c_{GradCAM}$ and $A^1_{ch}$. 

This process is summarized in Algorithm \ref{algo1}. As can be seen from the algorithm, the channels are combined by averaging them. Moreover, the algorithm performs pixel-wise inversion on the channels that show a negative correlation with the Grad-CAM heatmap. An alternative way of implementing this algorithm would be by using a computationally expensive and exhaustive approach that iterates through every single channel, every possible pair of channels, and so on for finding the best combination. We observed that the greedy approach achieves results that are very close to the results of the exhaustive approach in a more efficient way. 

\begin{figure}[htb]
  \centering
  \begin{minipage}{.6\linewidth}
    \begin{algorithm}[H]
    \caption{Combining channels of the first layer activations into one final visualization with high correlation with the GradCAM heatmap}
    \begin{algorithmic}[1]
    \State $A_1$: Array of first layer activations
    \State $L_{GCAM}$: Array of GradCAM heatmap
    \State $vis_{final}$: New array
    \State $ch_{idx}$: New list
    \State $vis_{corr}$: New list
    \State $corrs$: New list
    \While{$max(vis_{corr}) < max(corrs)$}
    	
    	\State $vis_{corr}.append(max(corrs))$
    	\State $idx \gets argmax(corrs)$
    	\State $ch_{idx}.append(idx)$
    	\State $vis_{final} \mathrel{+}= A_1[:,:, idx]$
    	\State $vis_{final} \mathrel{\slash}= (len(ch_{idx})+1) $

    	\For{$idx:=1$ to $n_{channels}$}
    		\State $vis \gets vis_{final}$
    		\State $vis \mathrel{+}= A_1[:,:,idx]$
    		\State $vis \mathrel{\slash}= (len(ch_{idx})+1) $
    		\State $corrs.append(\rho(L_{GCAM}, vis))$
    	\EndFor
    \EndWhile
    \State \textbf{return} $vis_{final}$
    \end{algorithmic}
    \label{algo1}
    \end{algorithm}
  \end{minipage}
\end{figure}

\subsubsection{Segmenting high-resolution visualizations}
\label{sec2-3-2}

Once a visualization/heatmap is created that highlights the burn regions in an image, it can be used for producing a segmentation mask. First, a Gaussian Mixture Model (GMM) is fitted to the pixel values of the generated visualization. Next, the points where Gaussian components meet are computed for the fitted model. We refer to these points as $\{t_i\}_{i=1}^{n_{components}}$. Finally, the heatmap is masked using these computed points in order to create a binary segmentation of the burn regions. For every threshold value $t_i$, the Intersection-Over-Union (IOU) score between the generated binary segmentation mask and the Grad-CAM heatmap is computed. The final threshold value (and therefore, the final binary mask) is selected to be the one that results in the highest IOU score. The generated binary segmentation mask then undergoes a post-processing step in order to filter out the noise regions. 

\subsection{Datasets}
\label{sec2-4}

\subsubsection{Burn Injury image dataset}
\label{sec2-4-1}

The primary dataset used for implementing the BAM method is a University of Alberta skin burn image dataset from clinics within Alberta Health Network. An REB approval (Pro00111990) was obtained for the purpose of training algorithms using de-identified patient images and data. The dataset contains a total of 1684 skin burn images taken using a standard digital camera. The burn severity of each image is labeled by burn surgeons. The labels include the four burn depth severities; SPF (superficial), SPT (superficial partial thickness), DPT (deep partial thickness), and FT (full thickness) burns. The number of images in each class is as follows; 243 SPF images, 799 SPT images, 463 DPT images, and 179 FT images. Pre-processing performed on the images include a CNN-based segmentation and the removal of background objects from images.  

\subsubsection{Laser Doppler Imaging dataset}
\label{sec2-4-2}

The LDI dataset includes a 2D colour image and a scan that shows the severity of burns and their complementary healing potential (HP) using a color palette. This smaller dataset consists of a total of 184 skin burn images and their associated LDI scans. The images of this smaller dataset belong to three different burn depth/degree classes as follows; 114 SPT images, 49 DPT images, and 21 FT images. The LDI scans were captured using the moorLDI laser Doppler imager (Moor Instruments Ltd) which is a non-invasive imaging device. 

In order for the LDI scans to be comparable with BAM binary segmentations, a number of processing steps were conducted. LDI scans can have different sizes, scales, and cropping in comparison to their corresponding burn images. As BAM uses the burn 2D colour images as the input for creating the binary burn segmentations, the LDI scans were first aligned with their corresponding burn images and converted into the same size as those images. Once the LDI scans are aligned with input images and their colors are processed in order to create binary masks, quantitative comparisons with BAM segmentations were conducted. For this purpose, we utilised the manual segmentations of burn areas from burn images validated by clinicians.

Moreover, as discussed later, it was discovered that the LDI scan color palette, which demonstrates different healing potentials, would classify uninjured areas and background noise in the image as burns with poor blood flow. In a clinical setting, this misclassification does not lead to a serious issue as scans are reviewed by clinicians who can easily differentiate between normal skin/background and burn area. However, since the processing of LDI scans is conducted by computer vision, this issue needed to be resolved. This was addressed by removing the non-burn areas from the LDI scans before processing LDI scans by multiplying the aligned LDI scans with the manual segmentations of burn areas resulting in LDI scans that show various healing potentials (or various degrees of burn) in the burn area only. 

\section{Results}
\label{sec3}

\begin{figure}[h]
     \centering
     \subfloat[]{\includegraphics[scale=0.405]{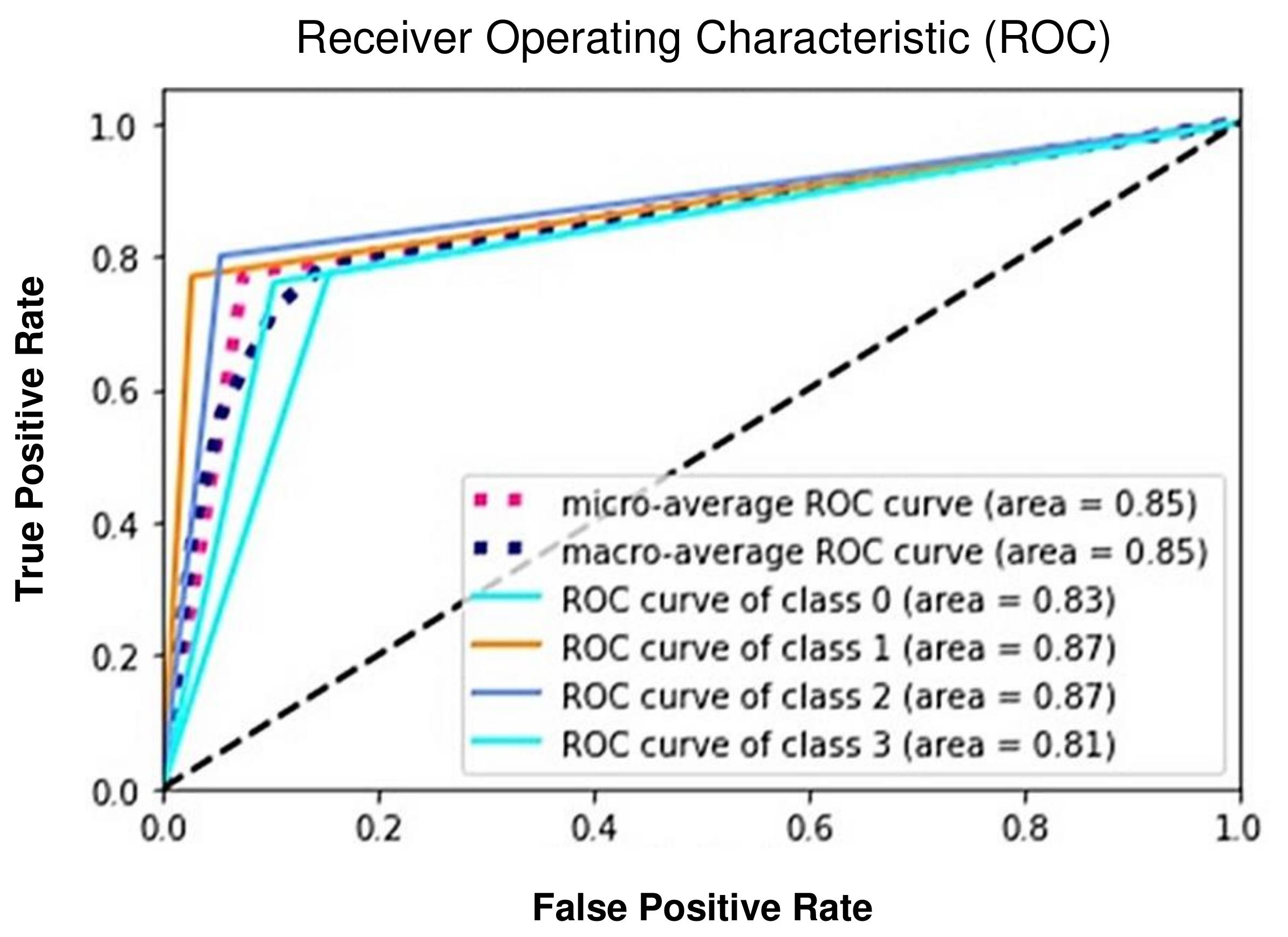}\label{fig1a}}
     \hspace{7pt}
     \subfloat[]{\includegraphics[trim={0 -70 0 -70}, scale=0.46]{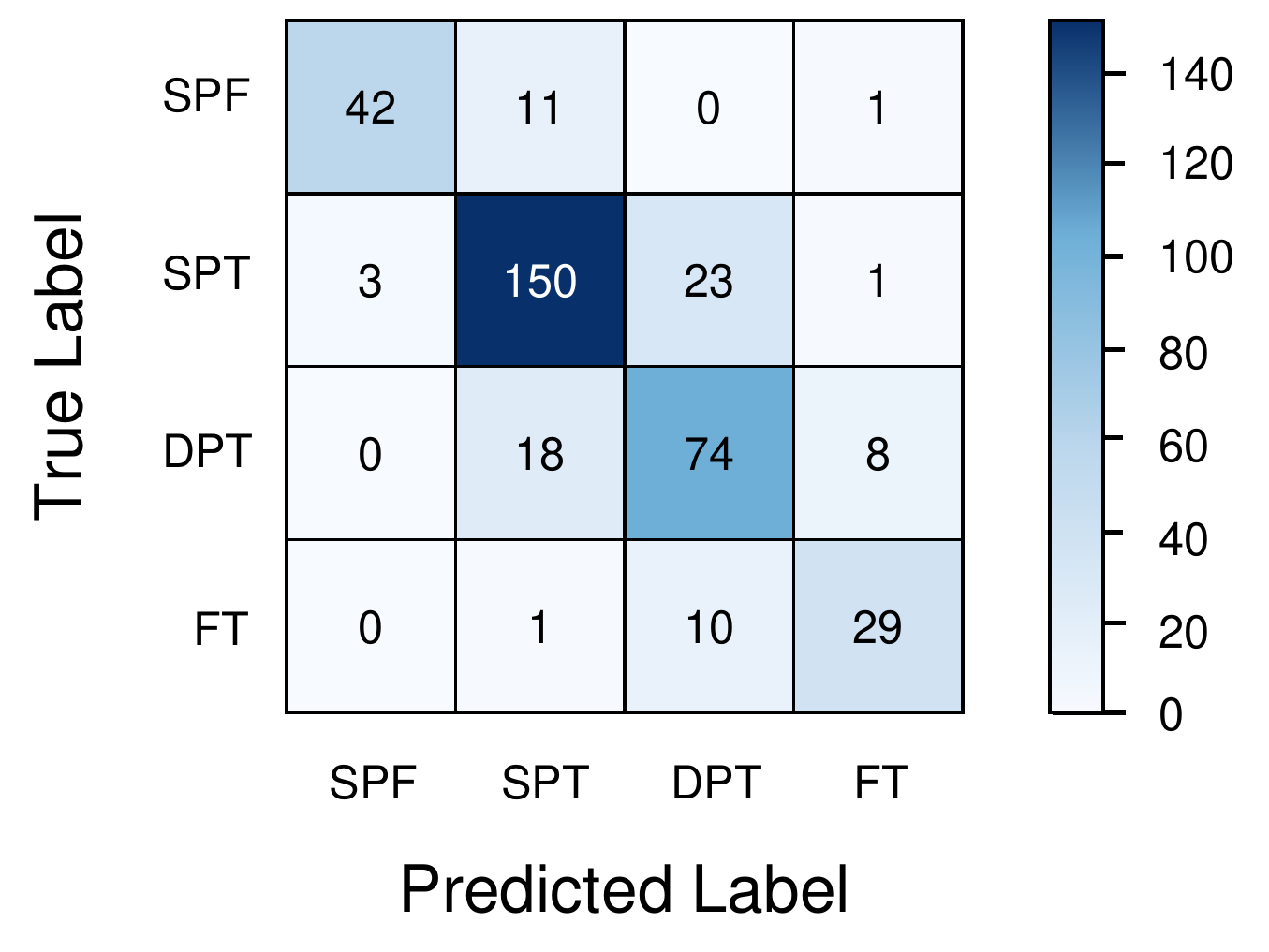}\label{fig1b}}
     \caption{a) ROC (receiver operating characteristics) curve and Area Under Curve (AUC) for the validation set computed for individual burn severity classes, and as micro- and macro- averages. b) Confusion matrix of the validation set computed for four individual burn severity classes.}
     \label{fig1}
\end{figure}

The Boundary Attention Mapper (BAM) methodology of creating saliency maps, dependent on information in various layers of a convolutional neural network (CNN), allows us to generate fine-grained segmentations of the burn injury from images. We trained a CNN, with a pre-trained EfficientNetB7 architecture, on 1684 burn images, to classify four severities of burns: SPF (superficial), SPT (superficial partial thickness), DPT (deep partial thickness), and FT (full thickness). The CNN achieved an average F1-Score of 78\% (Table \ref{table1}) and micro/macro- average ROC of 85\% (Figure \ref{fig1a}). A confusion matrix illustrates true and predicted values, therefore illustrating the true positive and negative error rates of the system. The matrix identifies that misclassification between burn severity classes is highest between adjacent classes of severity, for example SPT and DPT (Figure \ref{fig1b}). 

\begin{table}[h]
\begin{center}
\setlength{\tabcolsep}{10pt} 
\renewcommand{\arraystretch}{1.5} 
\begin{tabular}{ r | c c c }
\hline
\hline
\textbf{Burn Severity} & \textbf{Precision} & \textbf{Recall} & \textbf{F1-Score}\\ 
\hline
\hline
\textbf{SPF} & 0.93 & 0.78 & 0.85\\
\textbf{SPT} & 0.83 & 0.85 & 0.84\\
\textbf{DPT} & 0.69 & 0.74 & 0.71\\  
\textbf{FT} & 0.74 & 0.72 & 0.73\\
\hline
\textbf{Average} & 0.80 & 0.77 & 0.78\\
\hline
\hline
\end{tabular}
\caption{Precision, recall, and F1-score of the validation set computed using the trained CNN for classifying burn severities for each individual class and on average.}
\label{table1}
\end{center}
\end{table}

\begin{figure}[h]
     \centering
     \subfloat[]{\frame{\includegraphics[trim={-20 70 -20 -20}, width=0.495\textwidth]{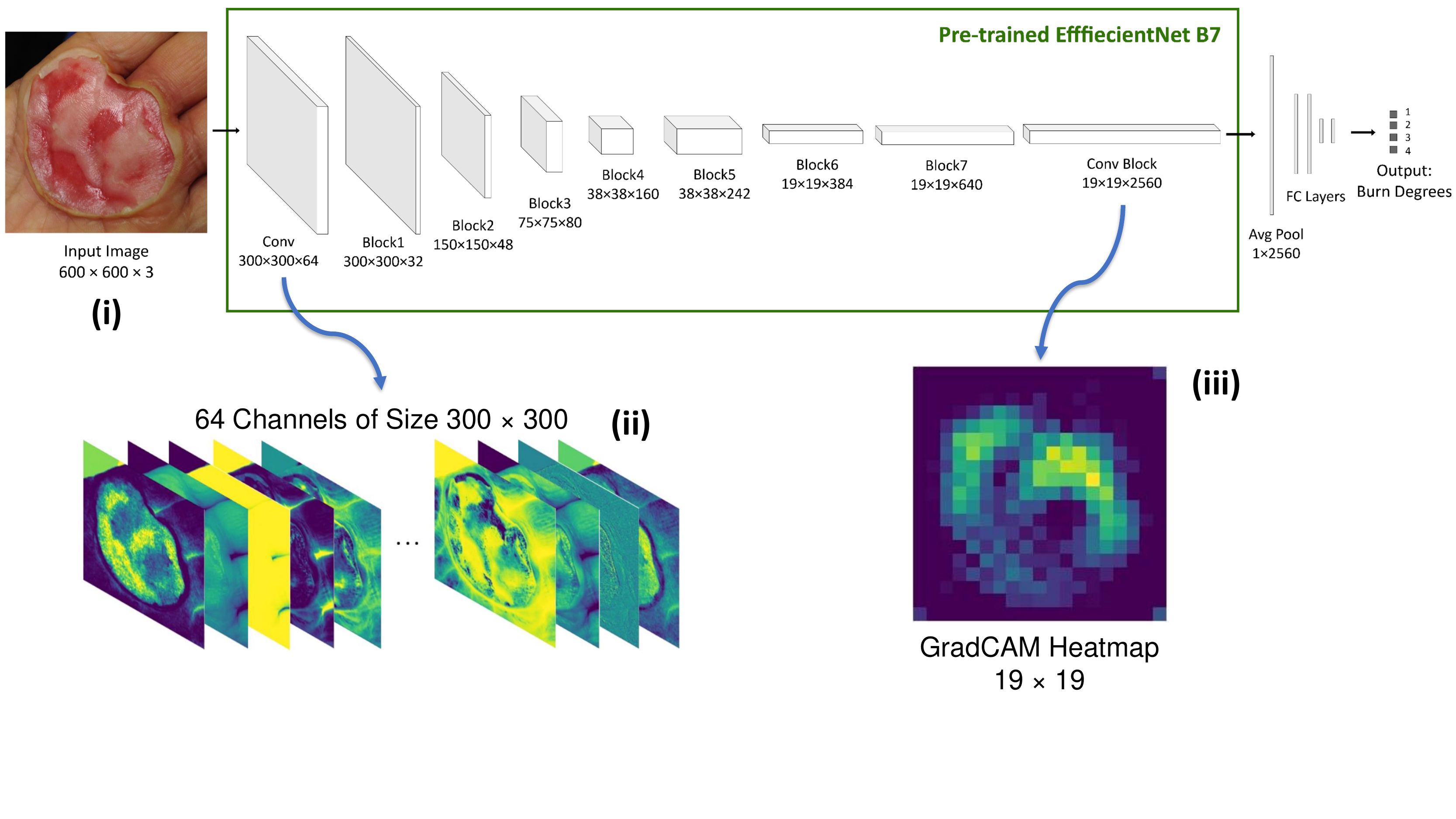}\label{fig2a}}}
     \hspace{5pt}
     \subfloat[]{\frame{\includegraphics[trim={-20 60 10 -20}, width=0.47\textwidth]{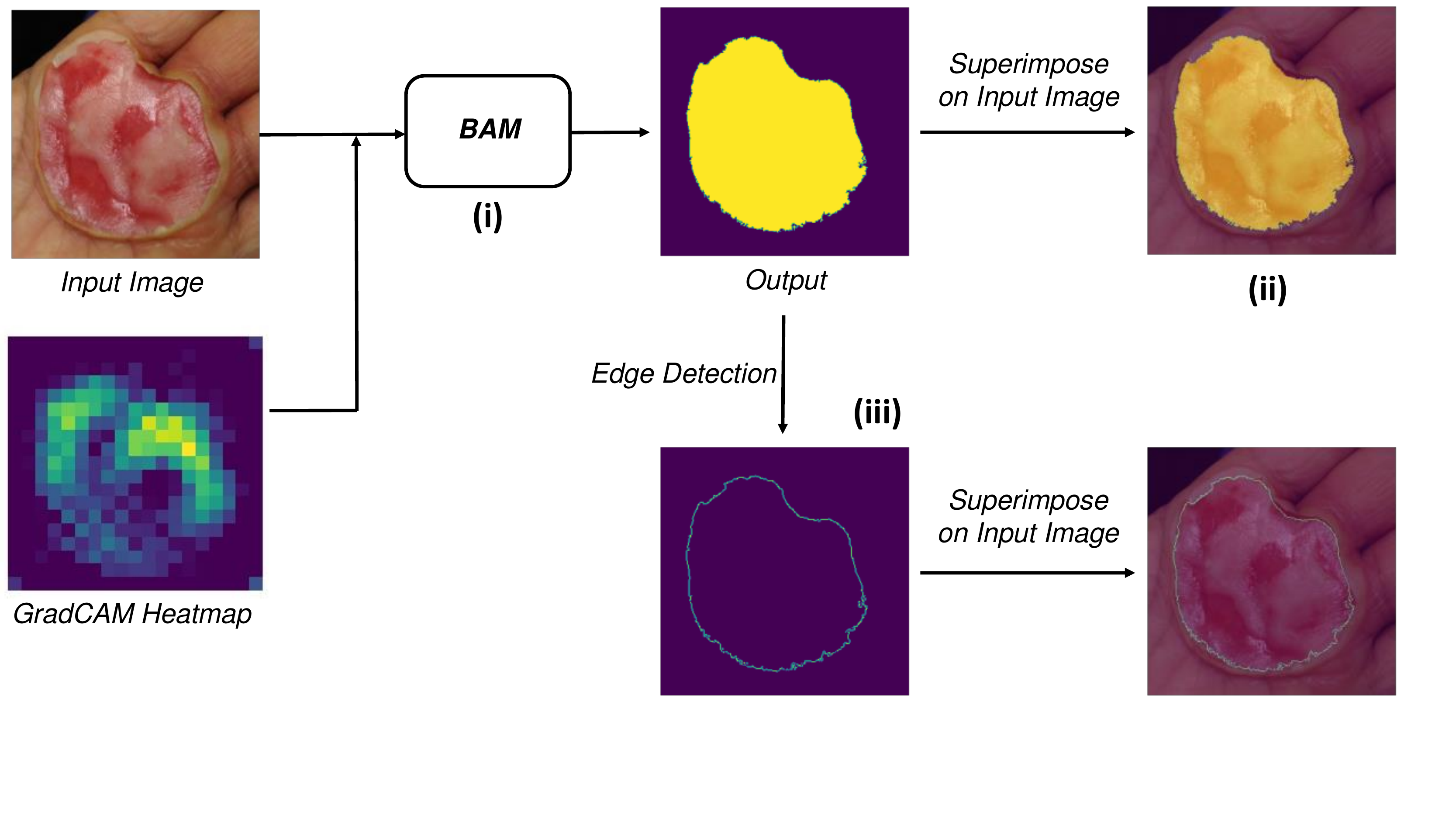}\label{fig2b}}}
     \caption{a) A schematic illustration of the transformations of a burn input 2D image in a feedforward pass of a CNN image classifier based on an EfficientNet architecture. An input image of a burn injury (i), the heatmaps for the activations of the first convolutional layer (ii), and the GradCAM heatmap computed using the last convolutional layer (iii) are depicted. b) An illustration of the inputs and the output of the Boundary Attention Mapping (BAM) method. Given an input skin burn image, the burn segmentation generated by BAM (i) is used to detect the area (ii) and the fine-tuned boundary (iii) of the burn. BAM can be used for measuring the burn area in pixels, or in absolute values using a fiducial marker (data not shown).}
     \label{fig2}
\end{figure}

\begin{figure}[h]
     \centering
     \subfloat{\includegraphics[trim={4 0 325 0}, width=0.475\textwidth]{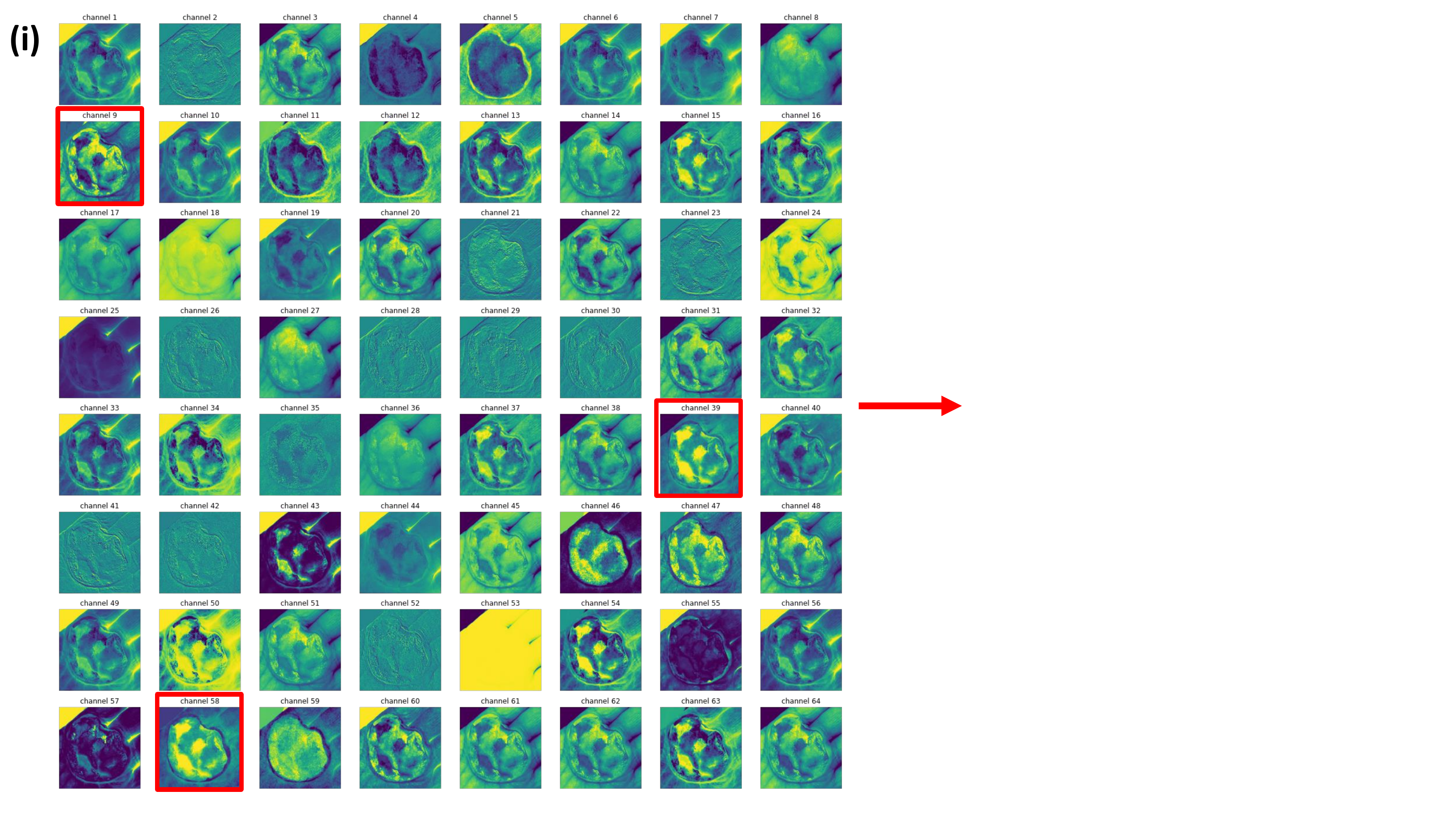}\label{fig3a}}
     \subfloat{\includegraphics[trim={0 -30 180 0}, width=0.53\textwidth]{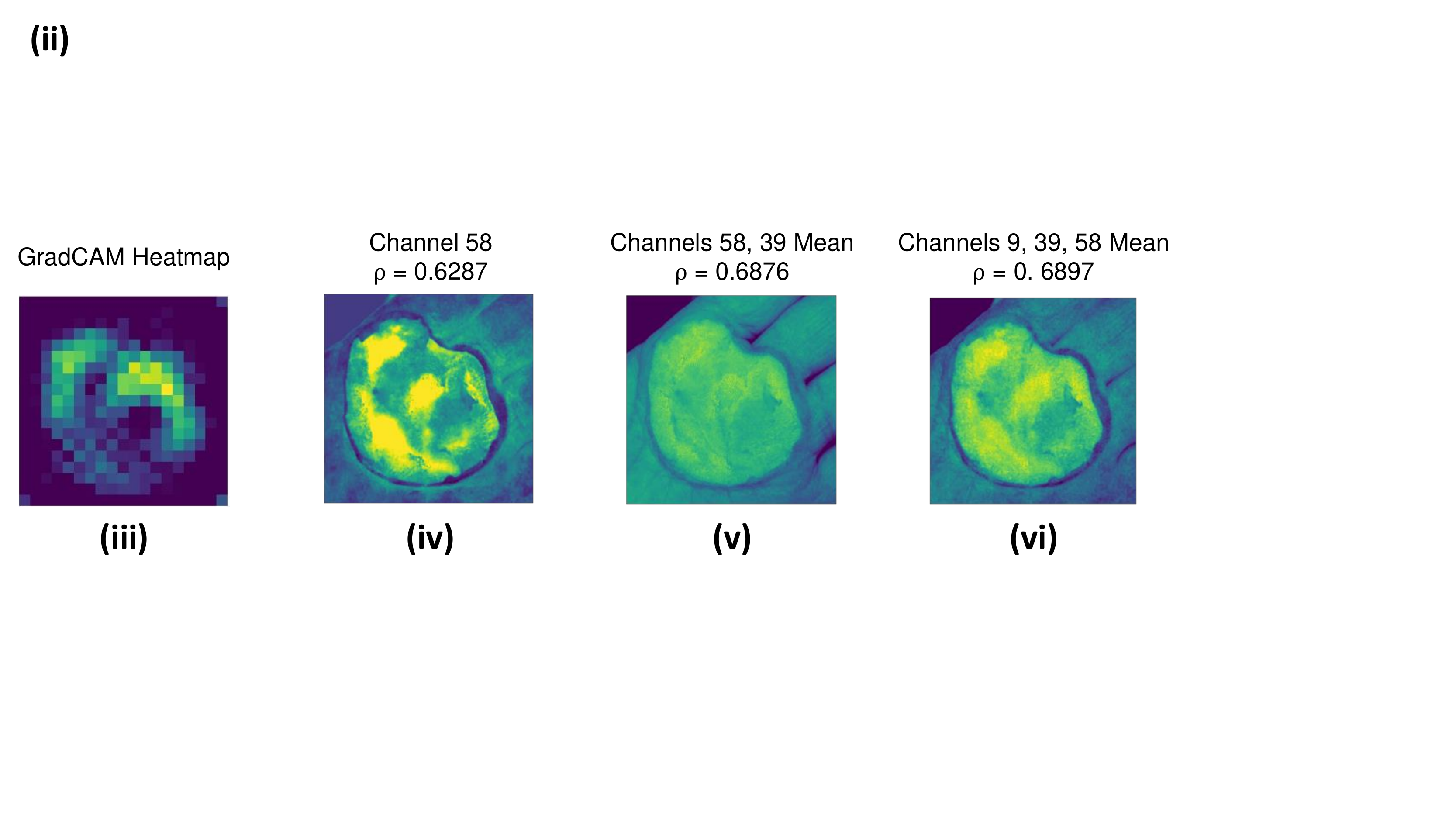}\label{fig3b}}
     \caption{(i) The heatmaps for each of the $64$ activation channels of the first convolutional layer from a burn image (see Figure \ref{fig2a}(ii). (ii) Iterations of selecting activation channels by Algorithm \ref{algo1} to produce a high-resolution visualization heatmap that is based on the highest correlation coefficient $\rho$ with the GradCAM heatmap. (iii) GradCAM heatmap, (iv) The first iteration of the algorithm selects activation channel $58$, (v) Iteration 2 of the algorithm selects channel $39$ to be averaged with channel $58$ in order to increase the correlation coefficient value $\rho$, (vi) Iteration 3 of the algorithm selects channel $9$ to be averaged with channels $58$ and $39$, in order to maximize the increase in the correlation between the activation-based map and the attention-based GradCAM.}
     \label{fig3}
\end{figure}

\begin{figure}[h]
	\centering
	\includegraphics[trim={0 -10 135 -10},clip, width=0.85\textwidth]{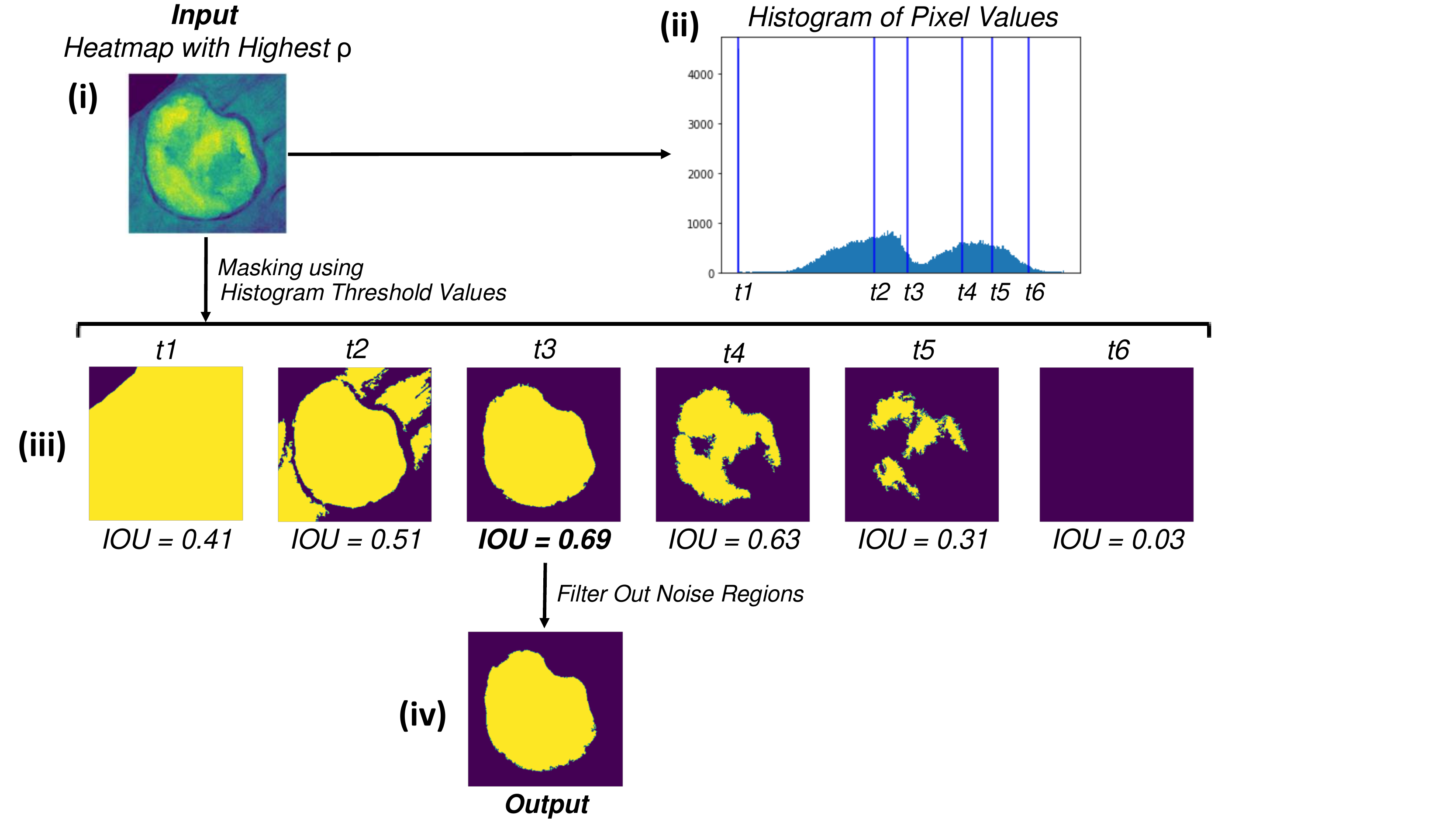}
	\caption{Masking the iterative heatmap (i), with the highest correlation with Grad-CAM, using threshold values from a histogram of pixel values (ii). Threshold values ($t_i$) equal the values where Gaussian components, fitted to the pixel values, intersect. For every threshold value $t_i$, the Intersection-Over-Union (IOU) score between the generated binary segmentation mask and the GradCAM heatmap is computed. The final binary mask selected has the highest IOU score ($t_3$, iii). The selected mask then undergoes a post-processing step in order to filter out noise (iv).}
	\label{fig4}
\end{figure}

Figure \ref{fig2a}. illustrates the information retrieved from this CNN model from various layers of the architecture that is used to create a BAM map, which is used to segment the burn injury from normal skin in a 2D image. First, the heatmaps for the activations of the first convolutional layer are computed (Figure \ref{fig2a}(ii)), and then Grad-CAM heatmap is computed using the last convolutional layer (Figure \ref{fig2a}(iii)). Once the first convolutional layer heatmaps and Grad-CAM are generated, the algorithm uses a three-round iterative process to select activation heatmaps that have the highest correlation to the Grad-CAM heatmap among the $64$ channels of the first layer activations. After the process of correlating and selecting heatmaps is completed (Figure \ref{fig3} ), segmentation masks are created next (Figure \ref{fig4}). A final composite BAM mask is created as illustrated in \ref{fig2b}(i). Finally, figure \ref{fig2b} (ii-iii) illustrates how the BAM mask is superimposed on the input image to segment the burn injury area, and how edge detection may be applied to the BAM mask in order to obtain a fine-tuned segmented boundary superimposed on the input image.

Figure \ref{fig3} and Algorithm \ref{algo1} detail the iterative process that is used to select the activation channel heatmaps and Grad-CAM, by the algorithm. For the example burn image, the first iteration of the algorithm selects channel $58$ since it has the highest correlation to the Grad-CAM heatmap among the $64$ channels of the first layer activations, as given by the Spearman's correlation coefficient, $\rho$. The second iteration selects channel $39$ to be added to the combination since averaging it with channel $58$ results in the highest increase in the $\rho$ correlation value with the Grad-CAM heatmap. Finally, the third iteration adds channel $9$ to the combination of channels $58$ and $39$. 

\begin{figure}[h]
\begin{center}
\subfloat{\includegraphics[trim=0 490 483 10, clip, width=0.475\textwidth]{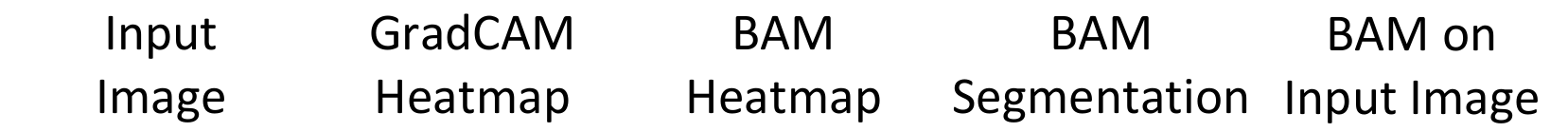}}
\hspace{10pt}
\subfloat{\includegraphics[trim=0 490 483 10, clip, width=0.475\textwidth]{figures/fig5/fig5_header}}
\\
\subfloat{\includegraphics[width=0.095\textwidth]{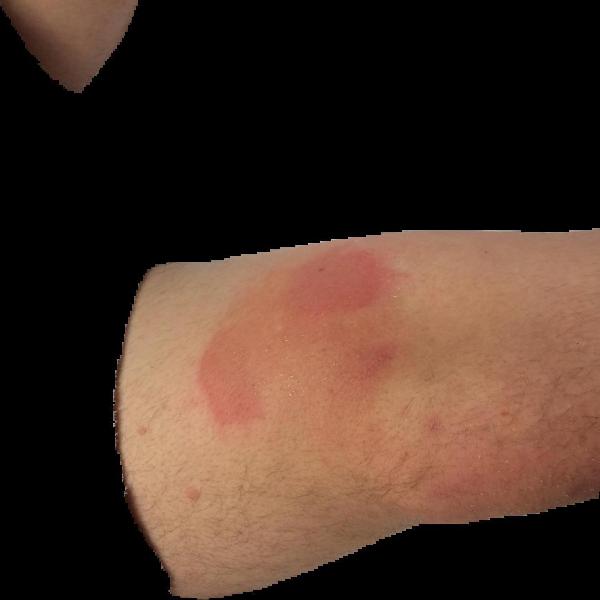}}
\subfloat{\includegraphics[width=0.095\textwidth]{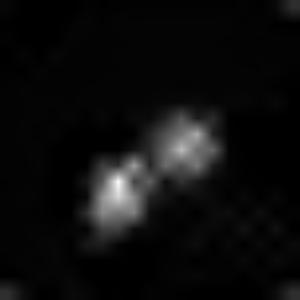}}
\subfloat{\includegraphics[width=0.095\textwidth]{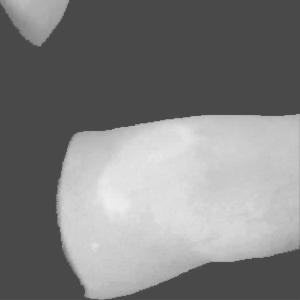}}
\subfloat{\includegraphics[width=0.095\textwidth]{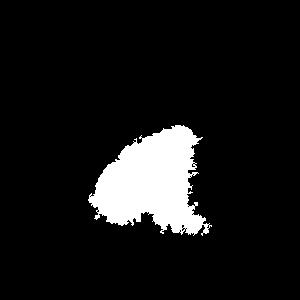}}
\subfloat{\includegraphics[trim=96 96 96 96, clip, width=0.095\textwidth]{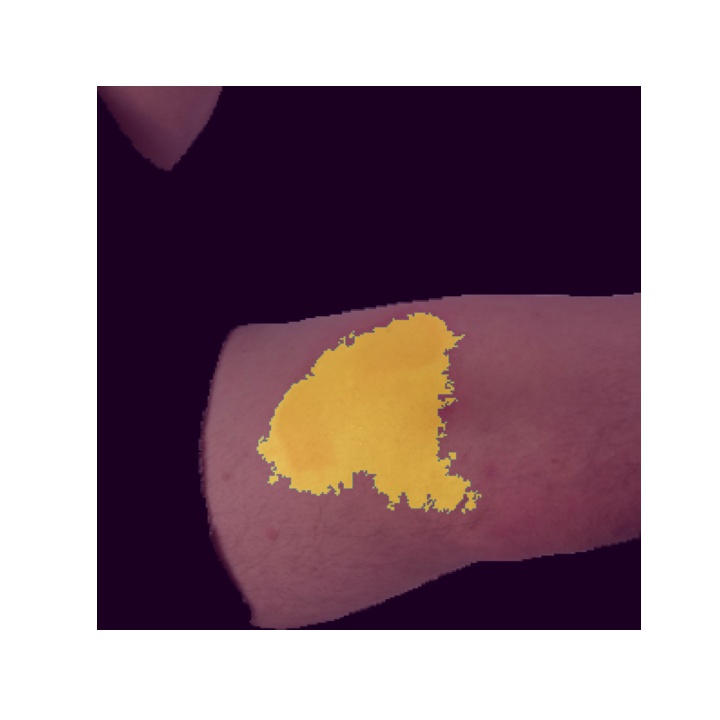}}
\hspace{10pt}
\subfloat{\includegraphics[width=0.095\textwidth]{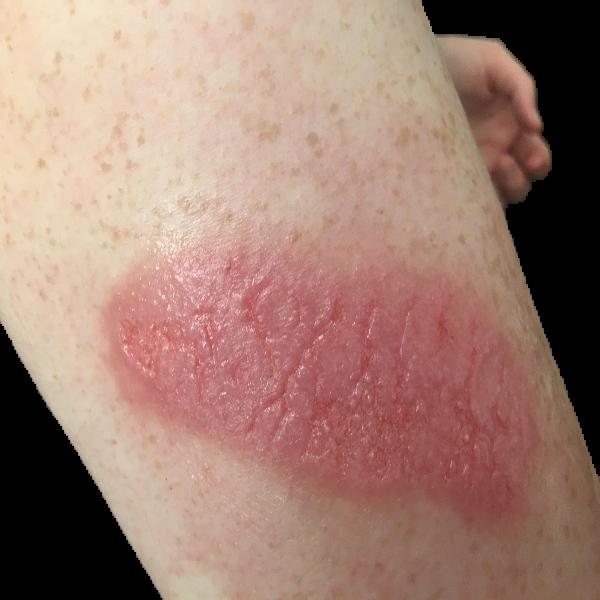}}
\subfloat{\includegraphics[width=0.095\textwidth]{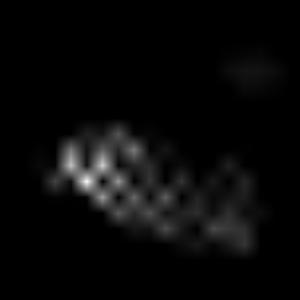}}
\subfloat{\includegraphics[width=0.095\textwidth]{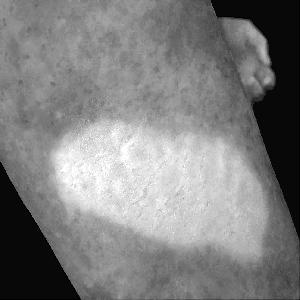}}
\subfloat{\includegraphics[width=0.095\textwidth]{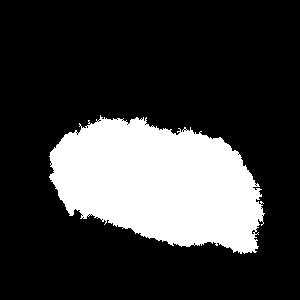}}
\subfloat{\includegraphics[trim=96 96 96 96, clip, width=0.095\textwidth]{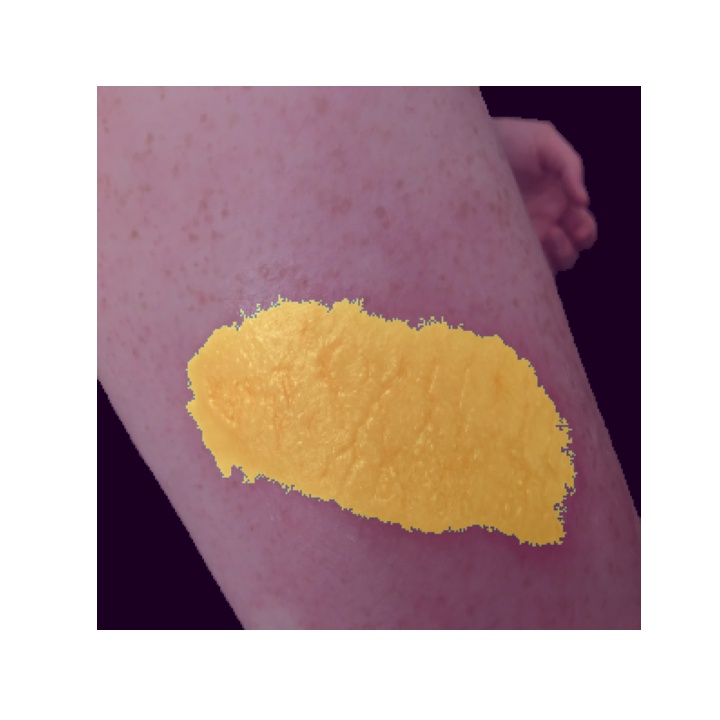}}
\\
\subfloat{\includegraphics[width=0.095\textwidth]{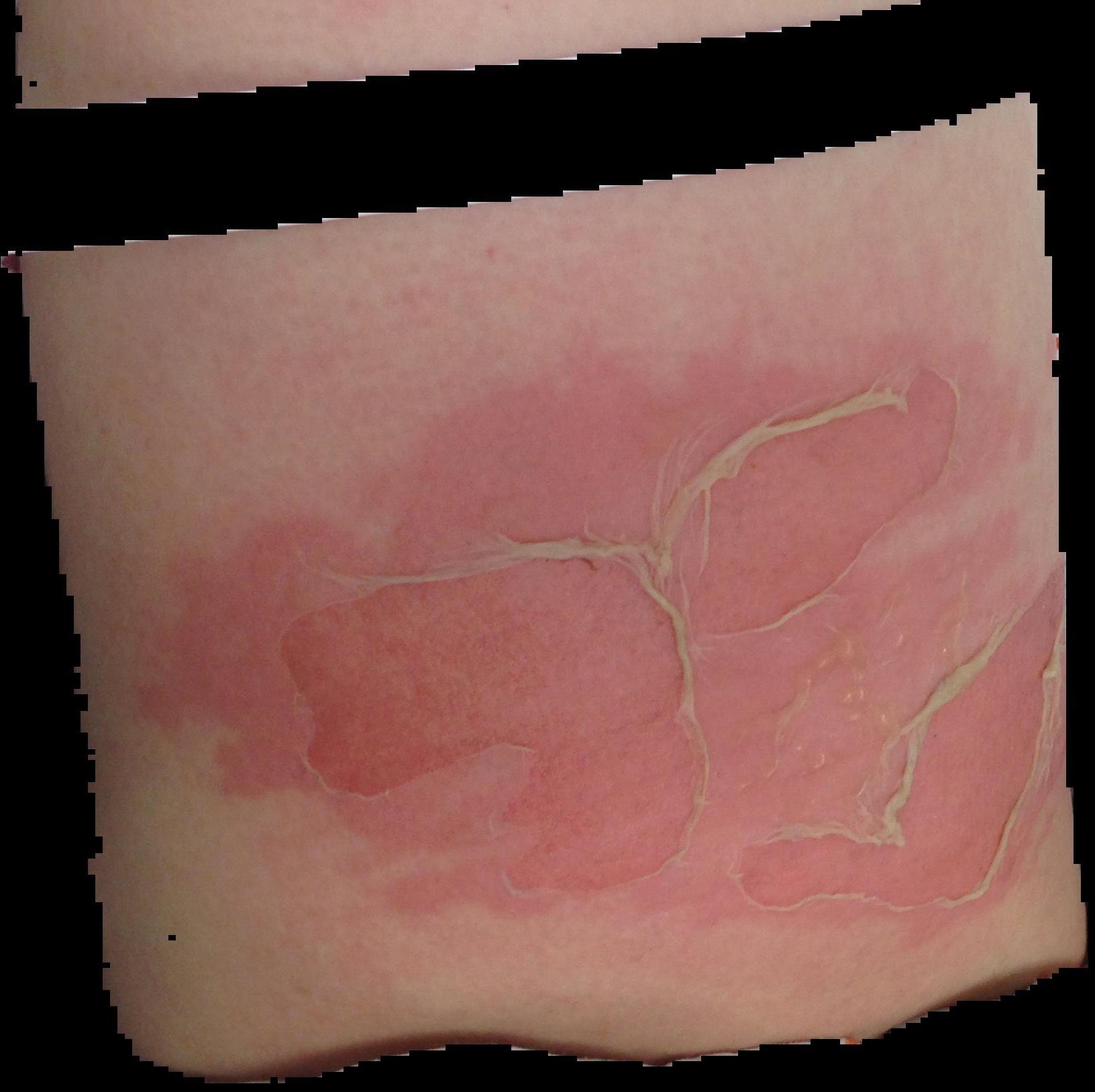}}
\subfloat{\includegraphics[width=0.095\textwidth]{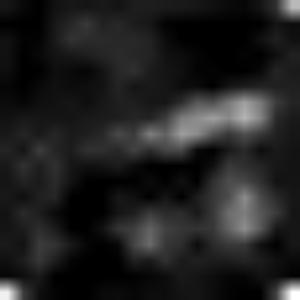}}
\subfloat{\includegraphics[width=0.095\textwidth]{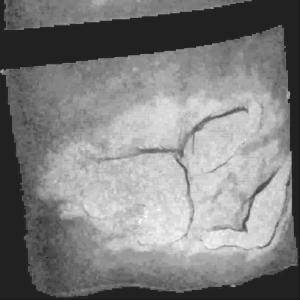}}
\subfloat{\includegraphics[width=0.095\textwidth]{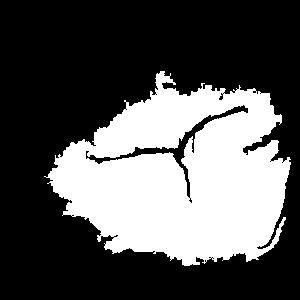}}
\subfloat{\includegraphics[trim=96 96 96 96, clip, width=0.095\textwidth]{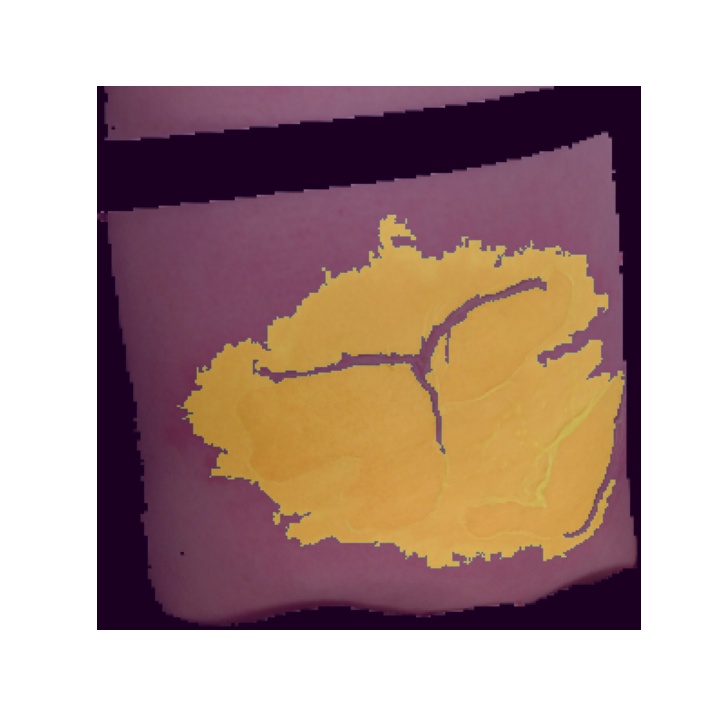}}
\hspace{10pt}
\subfloat{\includegraphics[width=0.095\textwidth]{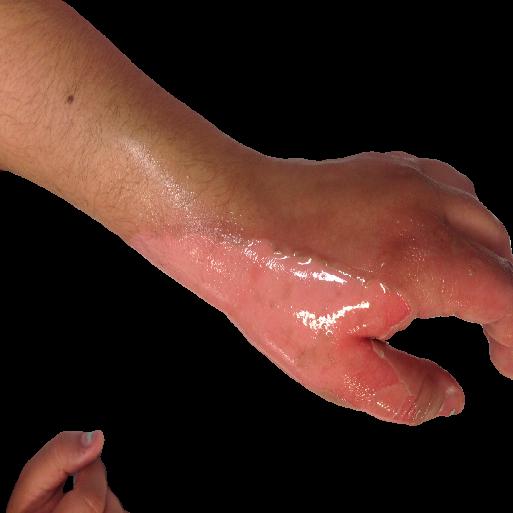}}
\subfloat{\includegraphics[width=0.095\textwidth]{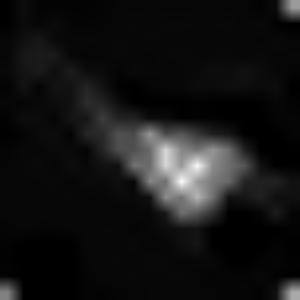}}
\subfloat{\includegraphics[width=0.095\textwidth]{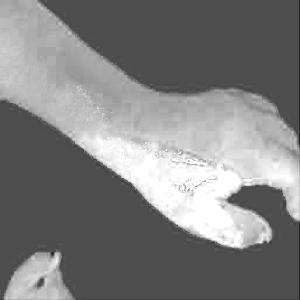}}
\subfloat{\includegraphics[width=0.095\textwidth]{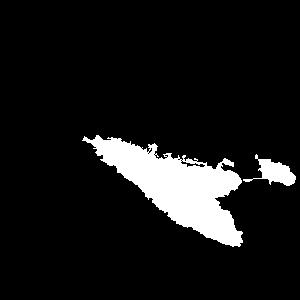}}
\subfloat{\includegraphics[trim=96 96 96 96, clip, width=0.095\textwidth]{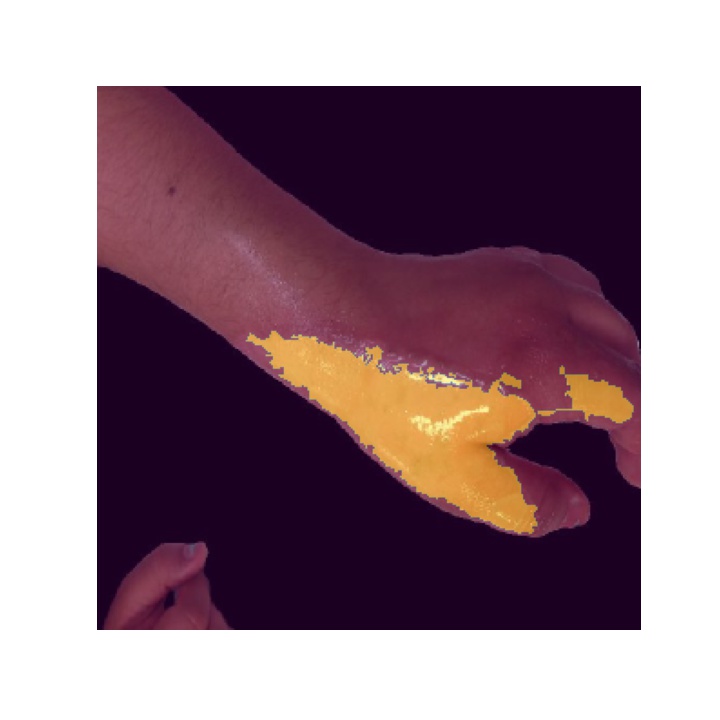}}
\\
\subfloat{\includegraphics[width=0.095\textwidth]{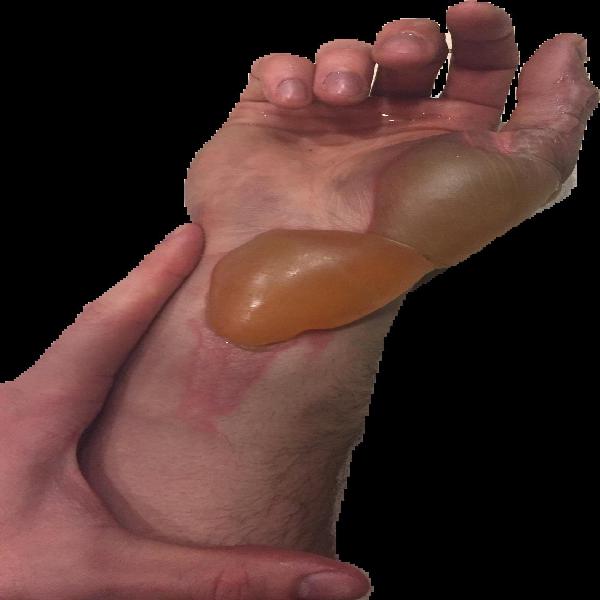}}
\subfloat{\includegraphics[width=0.095\textwidth]{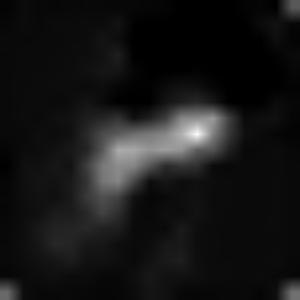}}
\subfloat{\includegraphics[width=0.095\textwidth]{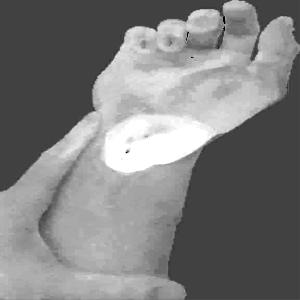}}
\subfloat{\includegraphics[width=0.095\textwidth]{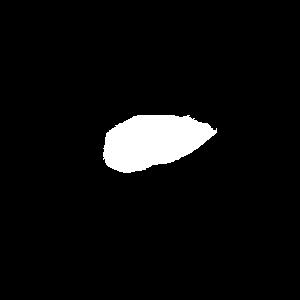}}
\subfloat{\includegraphics[trim=96 96 96 96, clip, width=0.095\textwidth]{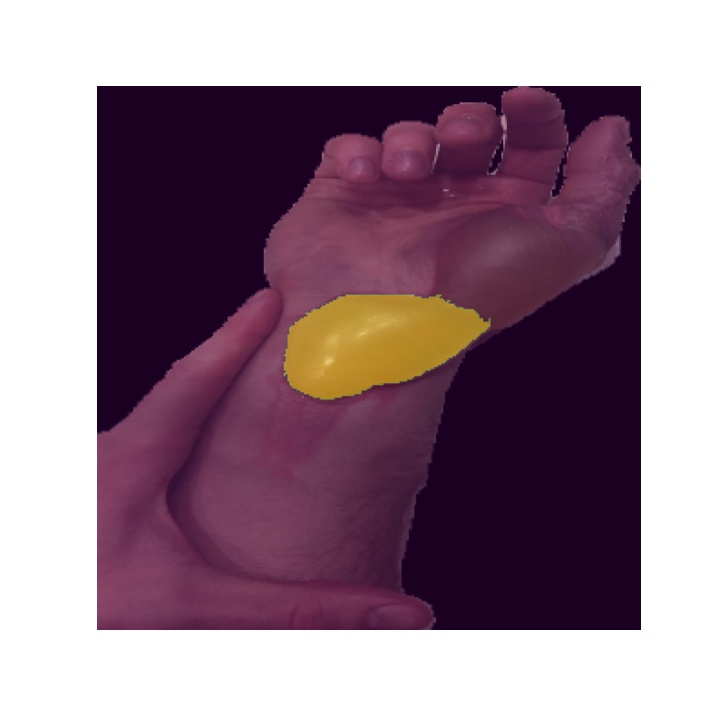}}
\hspace{10pt}
\subfloat{\includegraphics[width=0.095\textwidth]{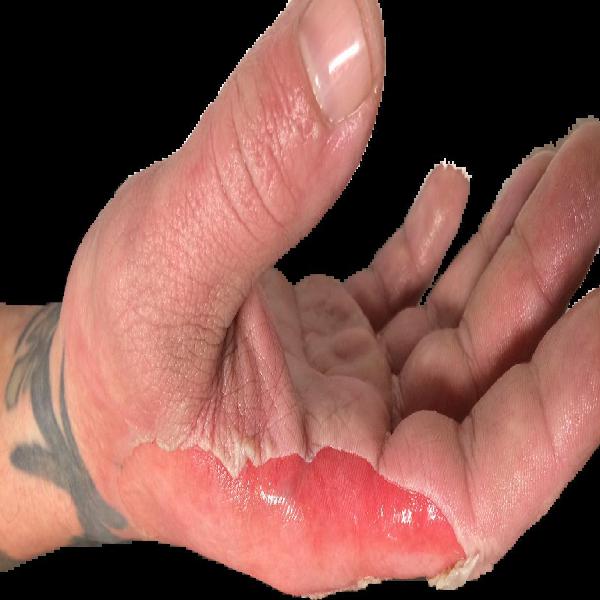}}
\subfloat{\includegraphics[width=0.095\textwidth]{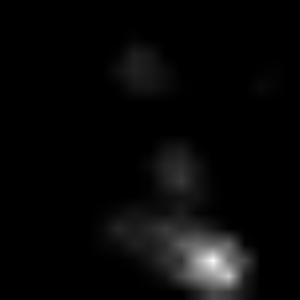}}
\subfloat{\includegraphics[width=0.095\textwidth]{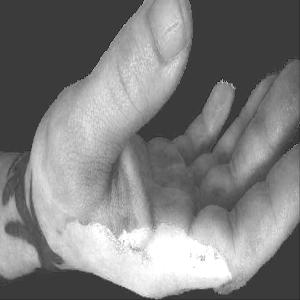}}
\subfloat{\includegraphics[width=0.095\textwidth]{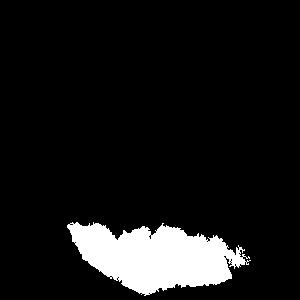}}
\subfloat{\includegraphics[trim=96 96 96 96, clip, width=0.095\textwidth]{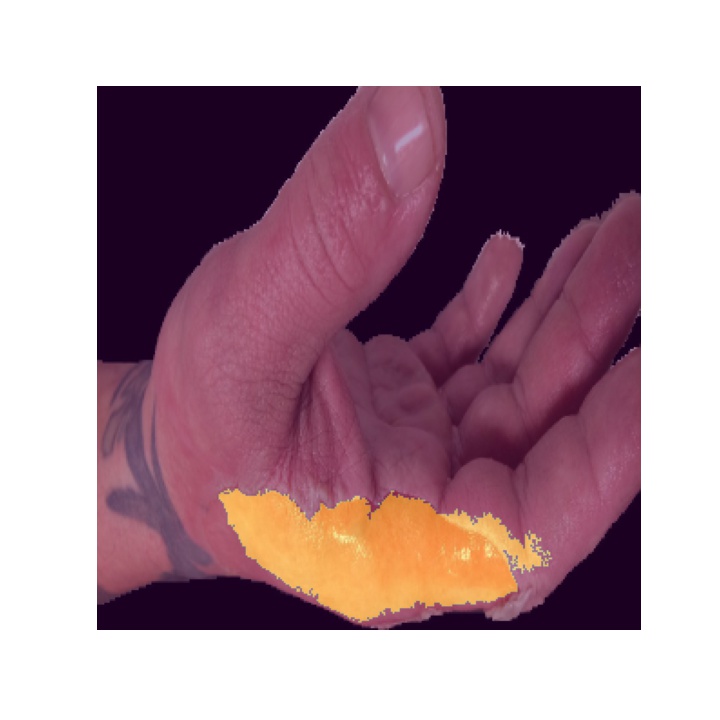}}
\\
\subfloat{\includegraphics[width=0.095\textwidth]{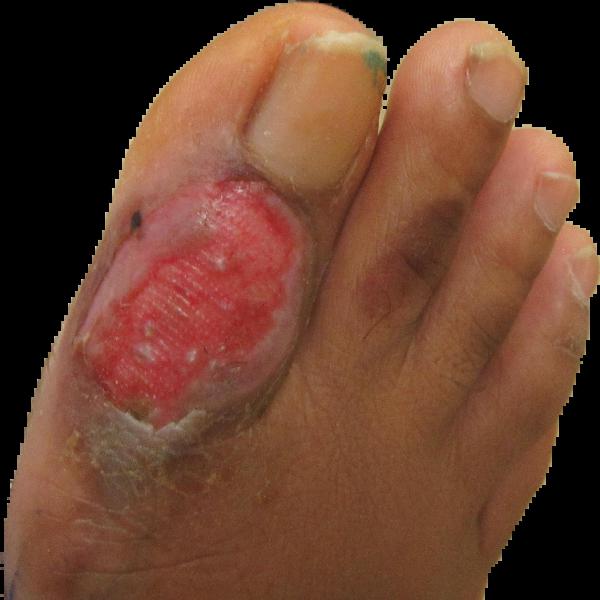}}
\subfloat{\includegraphics[width=0.095\textwidth]{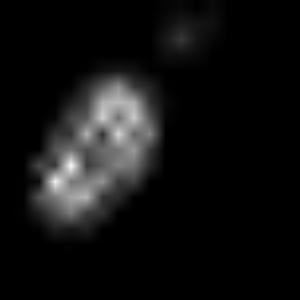}}
\subfloat{\includegraphics[width=0.095\textwidth]{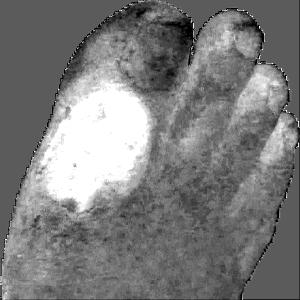}}
\subfloat{\includegraphics[width=0.095\textwidth]{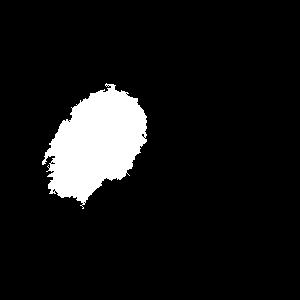}}
\subfloat{\includegraphics[trim=96 96 96 96, clip, width=0.095\textwidth]{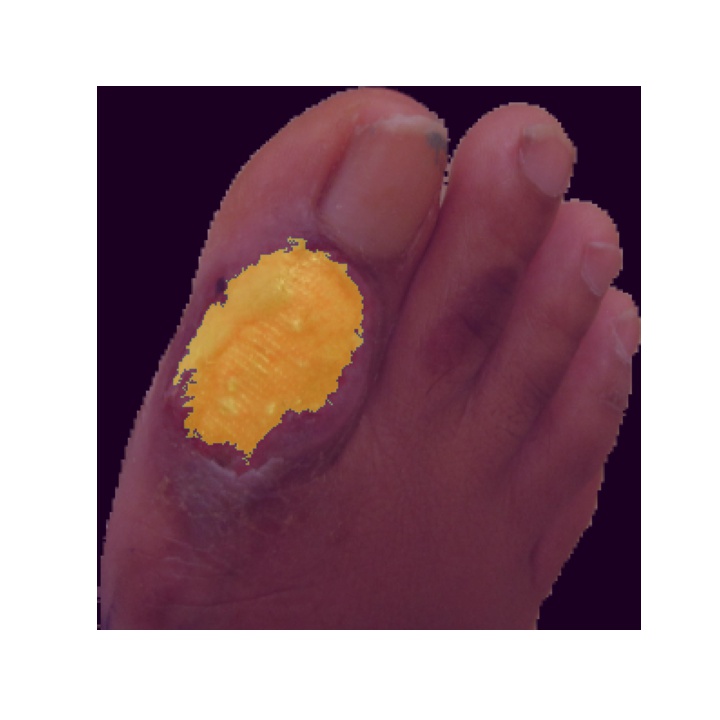}}
\hspace{10pt}
\subfloat{\includegraphics[width=0.095\textwidth]{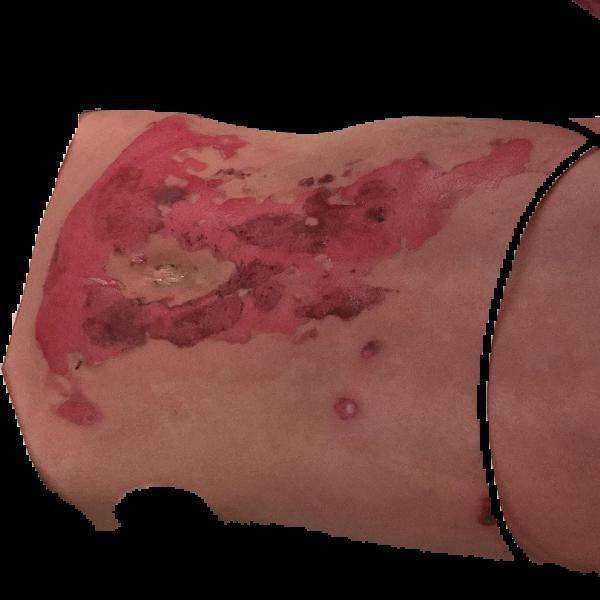}}
\subfloat{\includegraphics[width=0.095\textwidth]{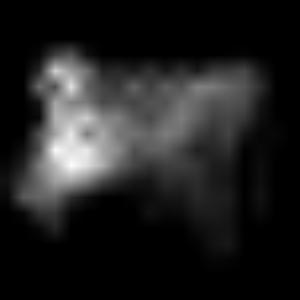}}
\subfloat{\includegraphics[width=0.095\textwidth]{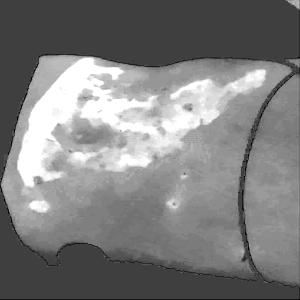}}
\subfloat{\includegraphics[width=0.095\textwidth]{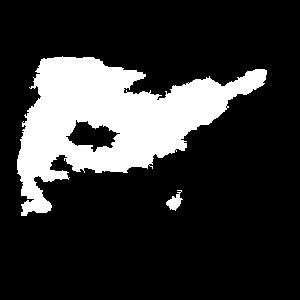}}
\subfloat{\includegraphics[trim=96 96 96 96, clip, width=0.095\textwidth]{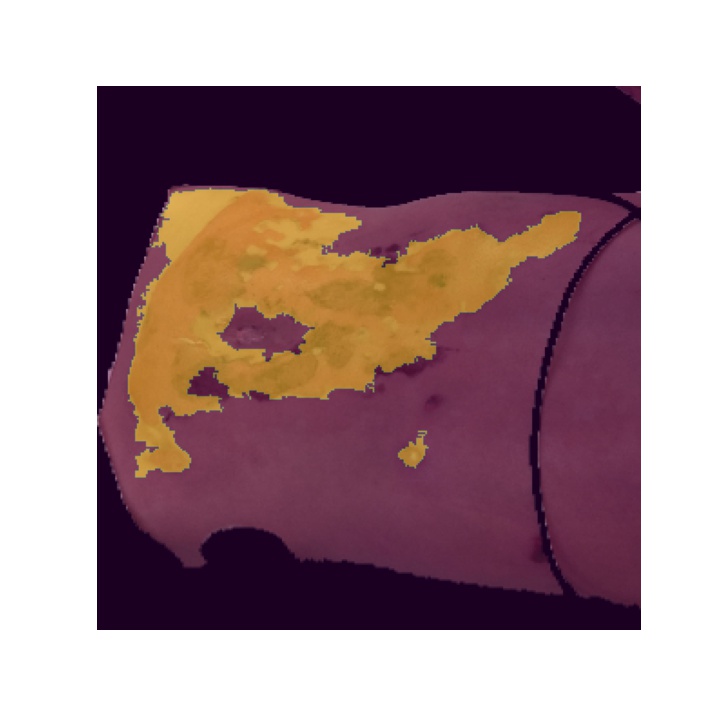}}
\\
\subfloat{\includegraphics[width=0.095\textwidth]{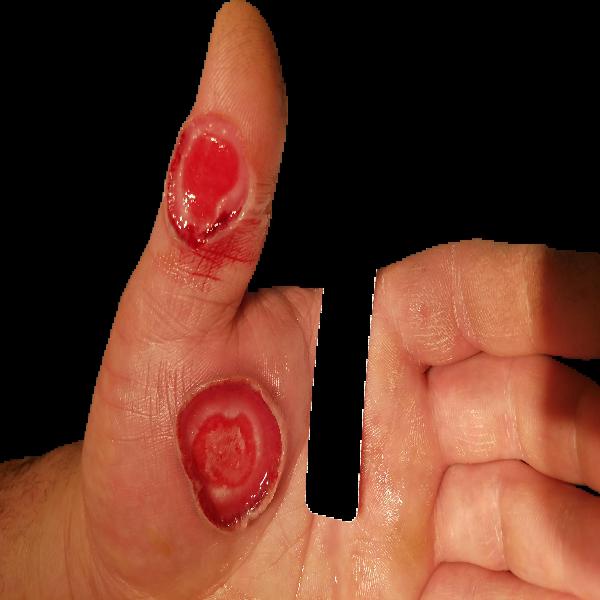}}
\subfloat{\includegraphics[width=0.095\textwidth]{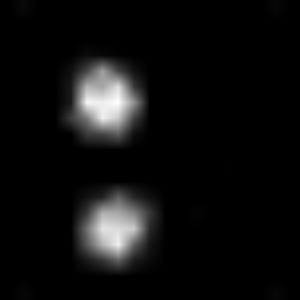}}
\subfloat{\includegraphics[width=0.095\textwidth]{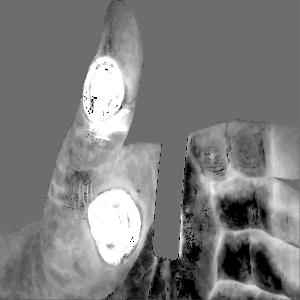}}
\subfloat{\includegraphics[width=0.095\textwidth]{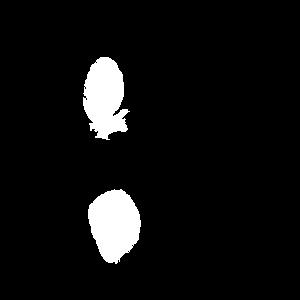}}
\subfloat{\includegraphics[trim=96 96 96 96, clip, width=0.095\textwidth]{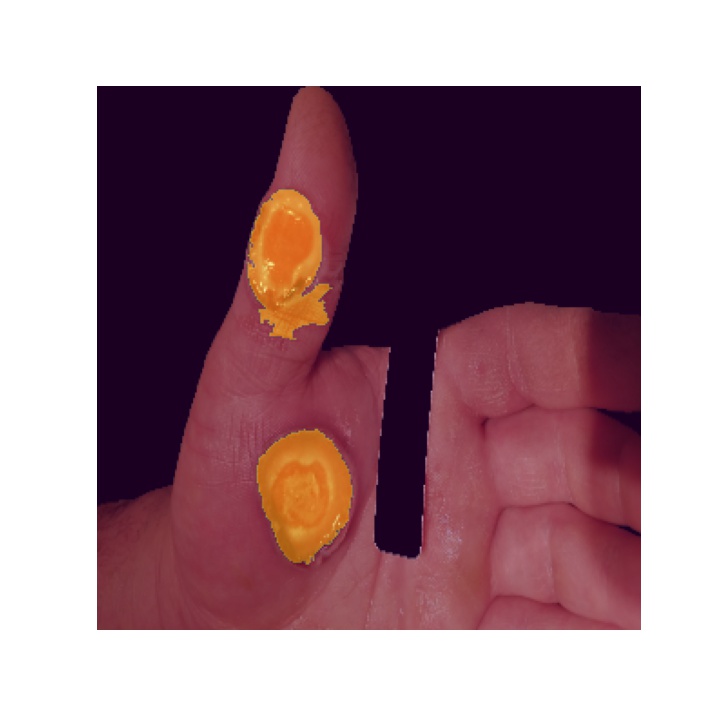}}
\hspace{10pt}
\subfloat{\includegraphics[width=0.095\textwidth]{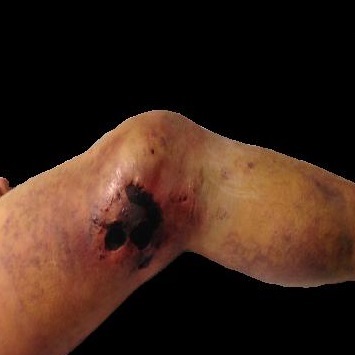}}
\subfloat{\includegraphics[width=0.095\textwidth]{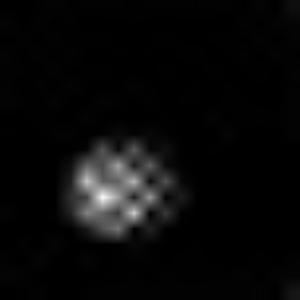}}
\subfloat{\includegraphics[width=0.095\textwidth]{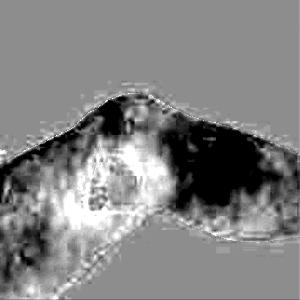}}
\subfloat{\includegraphics[width=0.095\textwidth]{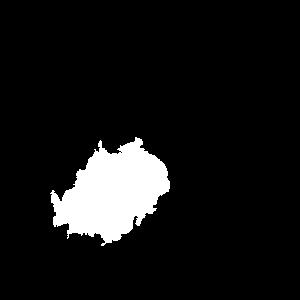}}
\subfloat{\includegraphics[trim=96 96 96 96, clip, width=0.095\textwidth]{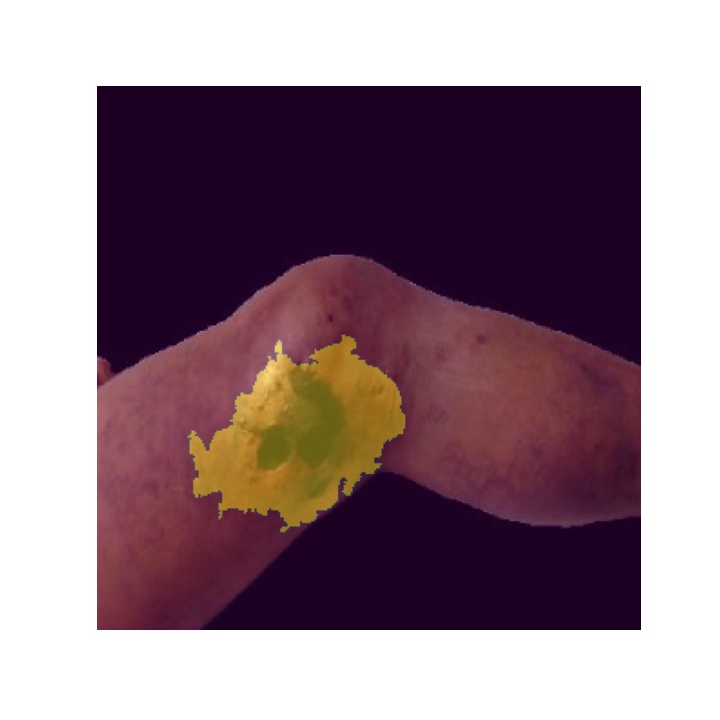}}
\\
\caption{Examples from the skin burn image dataset illustrating various degrees of burns. Each panel of images, from left to right, displays the following: a skin burn image, GradCAM heatmap, BAM heatmap, BAM segmentation, and the BAM segmentation super-imposed on the input image.}
\label{fig5}
\end{center}
\end{figure}

Once the heatmaps with the highest correlation coefficients are selected, these high-resolution visualizations are utilized as the input to make binary segmentation masks as illustrated in Figure \ref{fig4}(i). The generation of masks uses Gaussian components of the maps to find thresholds (Figure \ref{fig4}(ii)) and subsequently uses the highest Intersection-Over-Union (IOU) values (Figure \ref{fig4}(iii)) between the binary masks generated and the Grad-CAM to select the final mask. The generated binary segmentation mask lastly undergoes a post-processing step in order to filter out the noise/false positive regions and produce the final BAM mask (Figure \ref{fig4}(iv)), which can be used for super-positioning on the input image (Figure \ref{fig2b}). 

Figure \ref{fig5} shows several burn image examples of patients with different sized burns in different body locations, for which the Grad-CAM heatmap, BAM heatmap, BAM masks, and final superimposed images were created. These results allow us to understand the clinical accuracy of burn segmentation from 2D images using BAM. These images show various degrees of burn. It is evident from the results that given skin burn images and the corresponding Grad-CAM heatmaps highlighting the burn regions even partially, the BAM heatmap is able to highlight the burn regions and display a high resolution heatmap accurately. This is the main contribution of BAM. It can be seen from the figure that the BAM heatmaps display different contrast levels in highlighting the burn regions. More precisely, the more superficial burns are highlighted with a lower contrast to the normal skin. The deeper burns, on the other hand, are highlighted with a higher contrast to the normal skin. Nevertheless, the contrast between the burn regions and the normal skin in the BAM heatmaps is sufficient for generating the binary segmentation masks even for the more superficial burns. As evidenced, the BAM heatmaps can successfully be converted into accurate binary segmentation masks. The rightmost column of the figure shows the BAM segmentation masks on top of the input images in order to better visualize the effectiveness of BAM in segmenting the burn regions. In short, comparing the Grad-CAM heatmaps against the BAM heatmaps and BAM segmentation masks provides evidence for a significant improvement in generating heatmaps that are both class-discriminative and fine-grained.

\section{Quantitative Analysis}
\label{sec4}

\begin{figure}[h]
     \centering
     \subfloat[]{\frame{\includegraphics[trim={0 185 0 0}, width=0.95\textwidth]{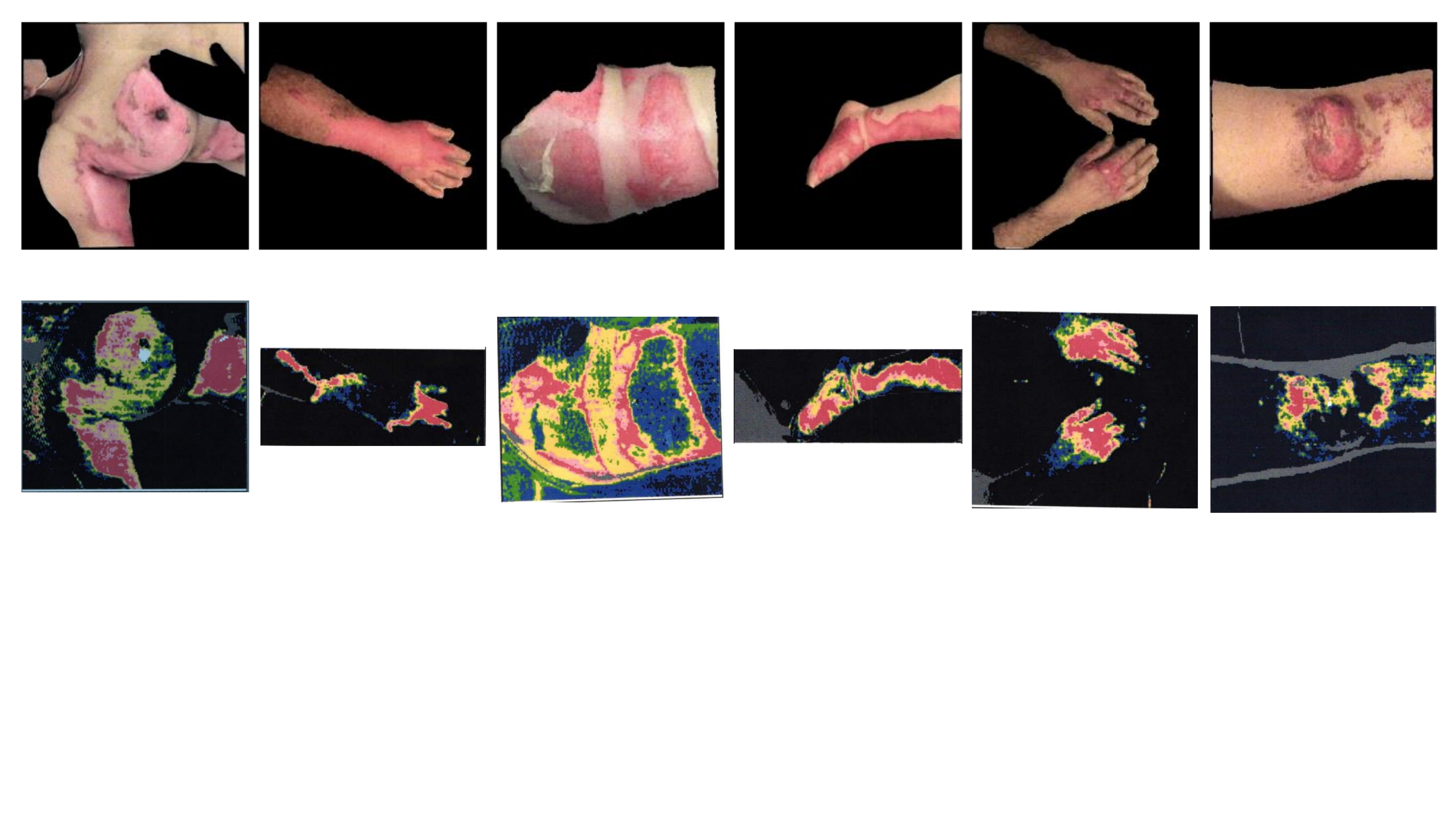}\label{fig6a}}}
     \\
     \subfloat[]{\includegraphics[trim={10 290 10 10}, width=0.96\textwidth]{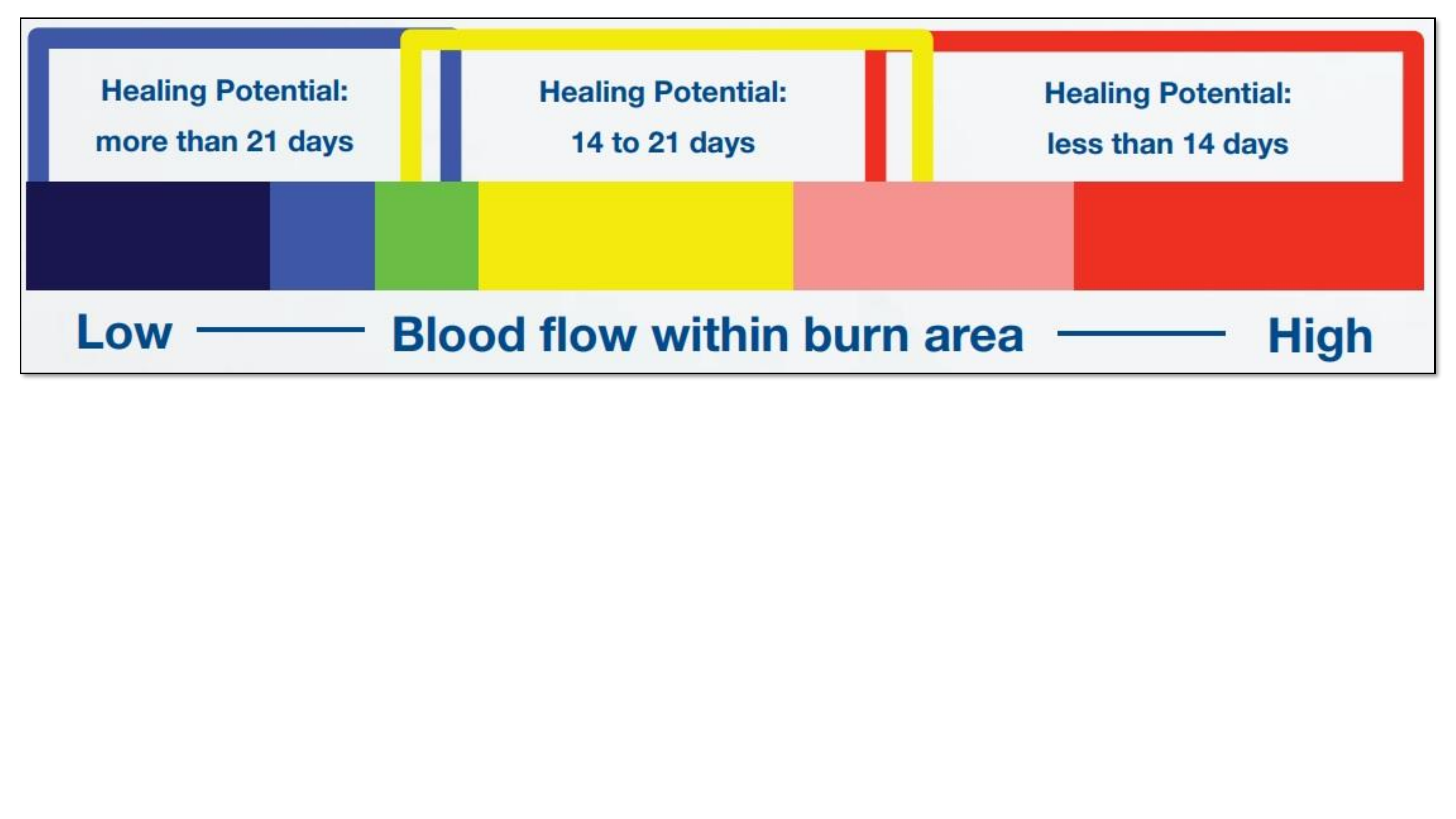}\label{fig6b}}
     \caption{a) Upper panel: Samples from the skin burn image dataset collected by The University of Alberta, Edmonton from clinics within Alberta Health Network, Canada. The images in the skin burn dataset have undergone a pre-processing step for the removal of the background noise. Lower Panel: The corresponding LDI scans for the images shown in the upper pannel. b) The colors of the blood flow image generated by moorLDI laser Doppler imager (Moor Instruments Ltd) and the three categories of healing potential for burn wounds represented in LDI scan colours}
     \label{fig6}
\end{figure}

We evaluated the performance of the BAM in segmenting burn areas from images using a dataset of manual segmentations validated by clinicians of burn areas from 2D colour images. We also compared BAM against Laser Doppler Imaging (LDI) results, the gold standard for assessing the depth and healing potential of burns. LDI generates a map of the blood flow in different parts of skin (including the burn areas) using laser Doppler technology. During scanning, laser light enters the skin tissue and is scattered by moving blood cells in the tissue. As a result, the frequency of the light changes according to the Doppler effect; the higher the speed and concentration of moving blood cells in a tissue, the higher the amplitude of the laser Doppler signal. This blood flow image is used to calculate three categories of healing potential for burn wounds; 1) less than 14 days, 2) 14 to 21 days, and 3) more than 21 days \cite{Medtech2021moorLDLS}\cite{hoeksema2009accuracy}. The colors of a blood flow image and their corresponding healing potential categories are illustrated in Figure \ref{fig6}. 

\begin{table}[H]
\begin{center}
\setlength{\tabcolsep}{10pt} 
\renewcommand{\arraystretch}{1.5} 
\begin{tabular}{ c r | c c c c }
\hline
\hline
& \textbf{Image Type} & \textbf{Accuracy} & \textbf{Sensitivity} & \textbf{Specificity} & \textbf{Jaccard-Index} \\
\hline
\hline
&&&&&\\
\multirow{5}{*}{\rotatebox[origin=c]{90}{\textbf{BAM}}} & \textbf{Manual Segmentations} & 93.33\% & 75.20\% & 96.47\% & 65.62\% \\
& \textbf{HP < 14} & 89.68\% & -- & 90.35\% & 36.80\% \\
& \textbf{14 < HP < 21} & 87.23\% & -- & 87.78\% & 19.49\% \\
& \textbf{HP > 21} & 86.62\% & -- & 86.83\% & 9.21\% \\
& \textbf{LDI (all HPs)} & 91.60\%  & 78.17\% & 93.37\% & 53.71\% \\
&&&&&\\
\hline
\multirow{4}{*}{\rotatebox[origin=c]{90}{\textbf{GradCAM}}}&&&&&\\
  & \textbf{Manual Segmentations} & 81.73\% & 44.18\% & 94.58\% & 37.20\% \\
& \textbf{LDI (all HPs)} & 83.78\%  & 48.29\% & 92.17\% & 31.65\% \\
&&&&&\\
\hline
\hline
\end{tabular}
\caption{Top: A comparison of pixel-wise accuracy, pixel-wise sensitivity, pixel-wise specificity, and Jaccard-Index between the BAM segmentations and 1) Manual segmentations, 2) LDI \textit{HP < 14 days}, 3) LDI \textit{14 days < HP < 21 days}, 4) LDI \textit{HP > 21} days, 5) LDI all three HPs. Bottom: A comparison of pixel-wise accuracy, pixel-wise sensitivity, pixel-wise specificity, and Jaccard-Index between the GradCAM segmentations and 1) Manual segmentations, 2) LDI all three HPs. }
\label{table2}
\end{center}
\end{table}


\begin{figure}[h]
	\centering
	\includegraphics[trim={-5 -5 115 0},clip, width=0.9\textwidth]{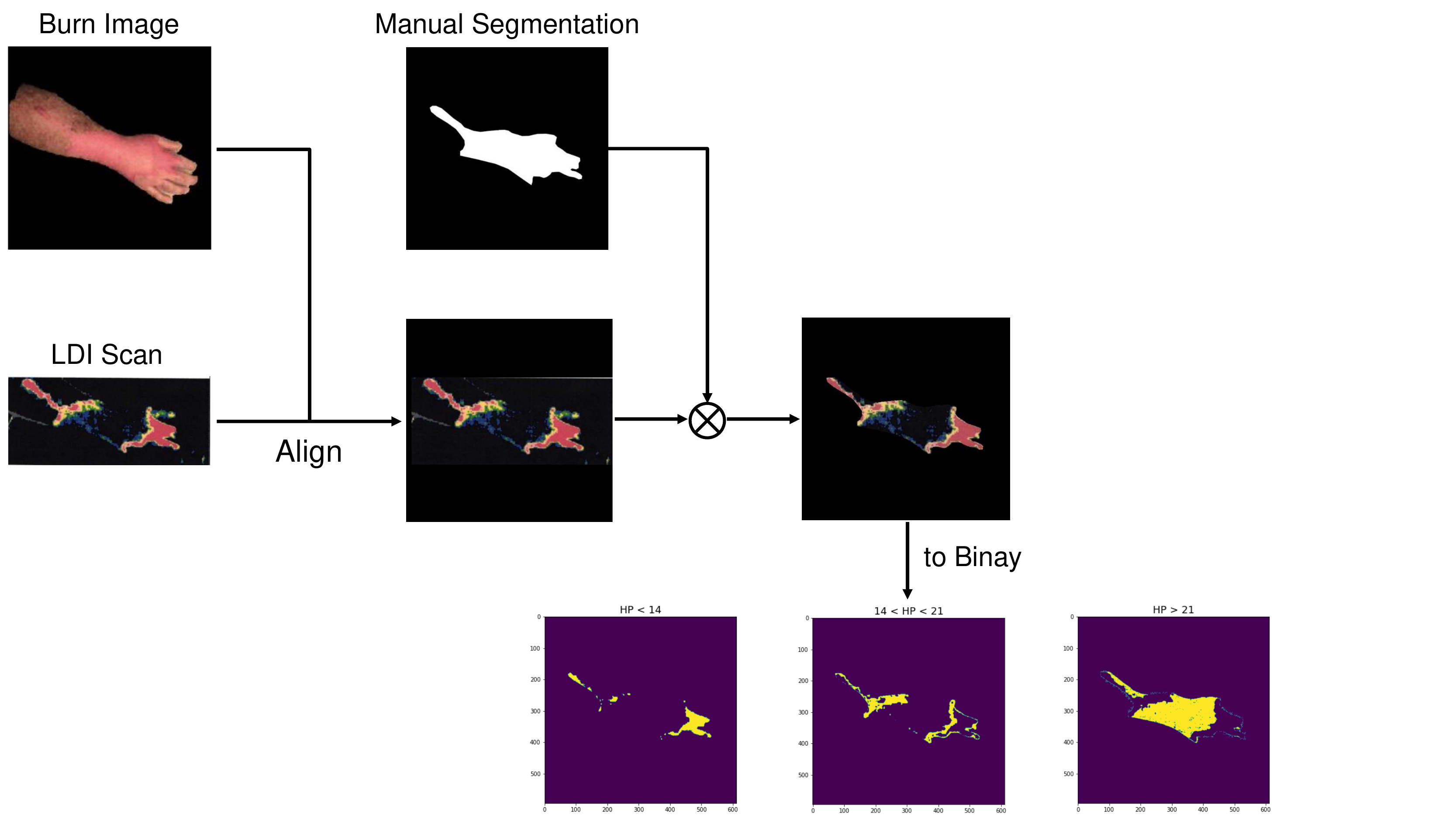}
	\caption{Processing of Laser Doppler Imaging (LDI) scans in order to create a benchmark dataset for evaluation of BAM burn segmentation methodology.}
	\label{fig7}
\end{figure}

\begin{figure}[h]
\begin{center}
\subfloat{\includegraphics[trim={250 471 265 25},clip, width=0.915\textwidth]{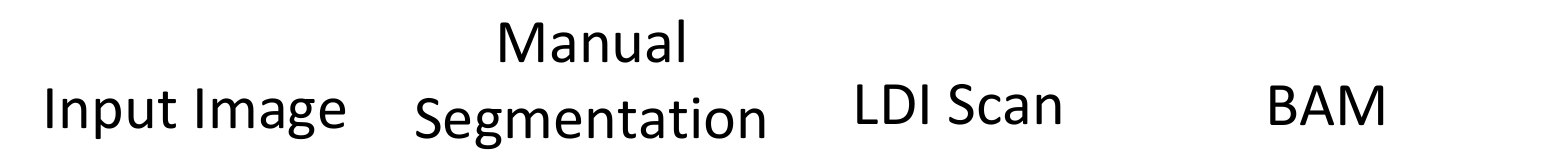}}
\\
\subfloat{\includegraphics[width=0.22\textwidth]{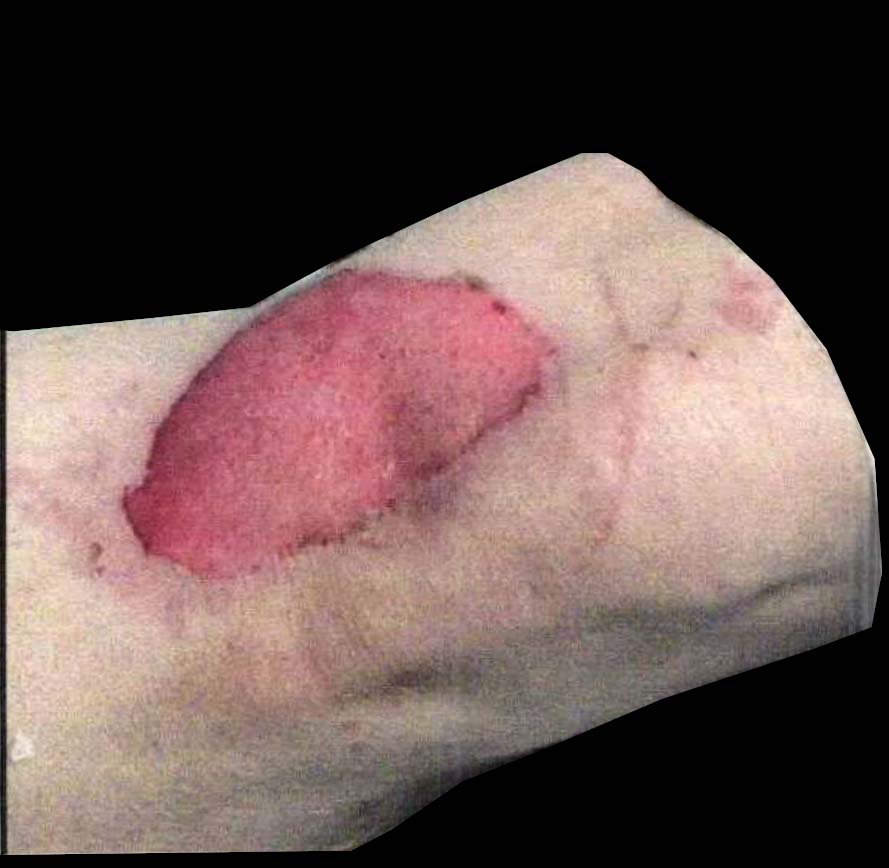}}
\hspace{1pt}
\subfloat{\includegraphics[width=0.22\textwidth]{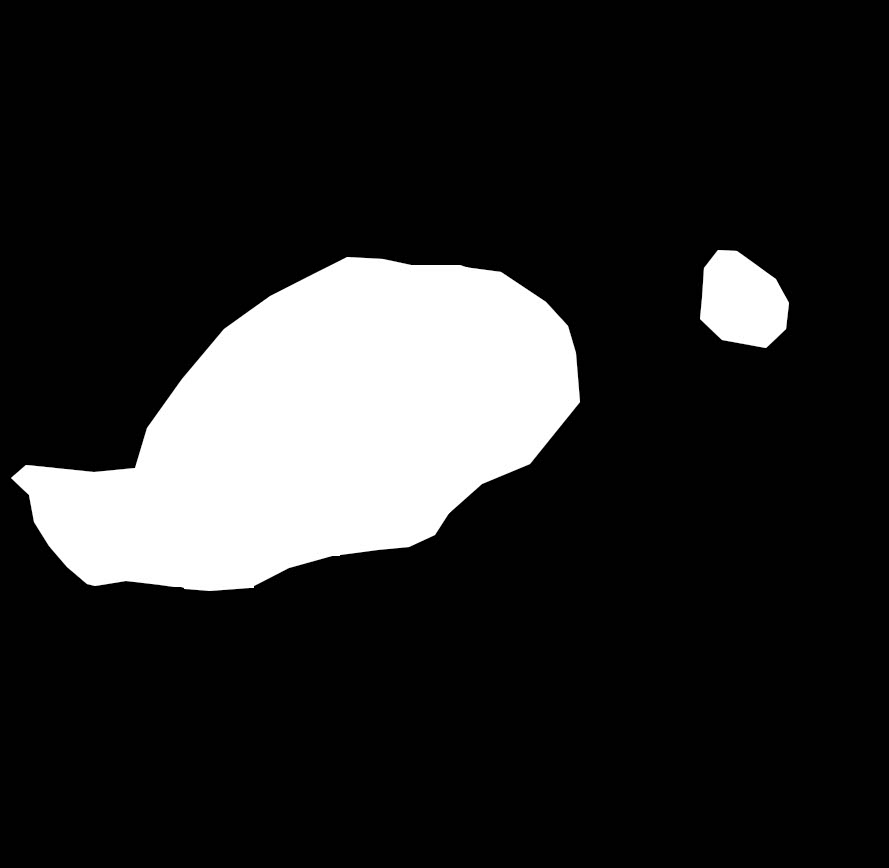}}
\hspace{1pt}
\subfloat{\includegraphics[width=0.22\textwidth]{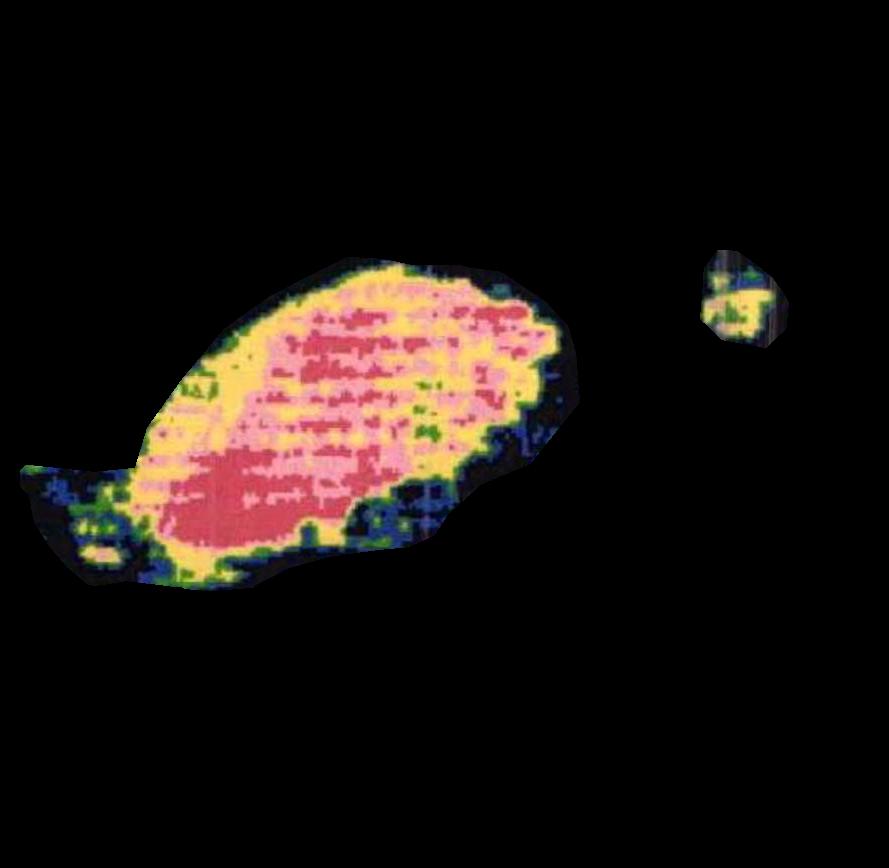}}
\hspace{1pt}
\subfloat{\includegraphics[width=0.215\textwidth]{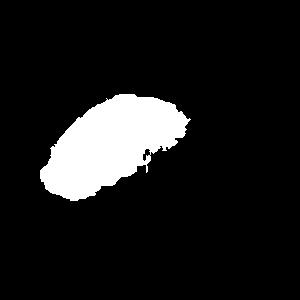}}
\\
\subfloat{\includegraphics[width=0.22\textwidth]{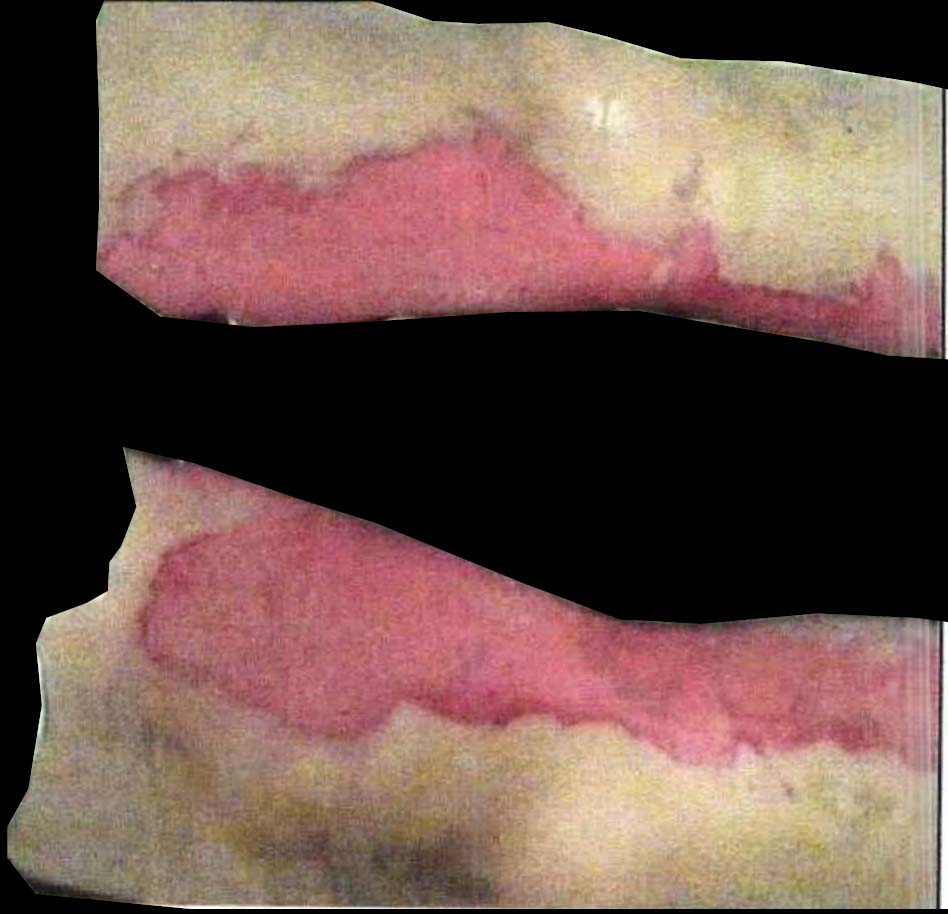}}
\hspace{1pt}
\subfloat{\includegraphics[width=0.22\textwidth]{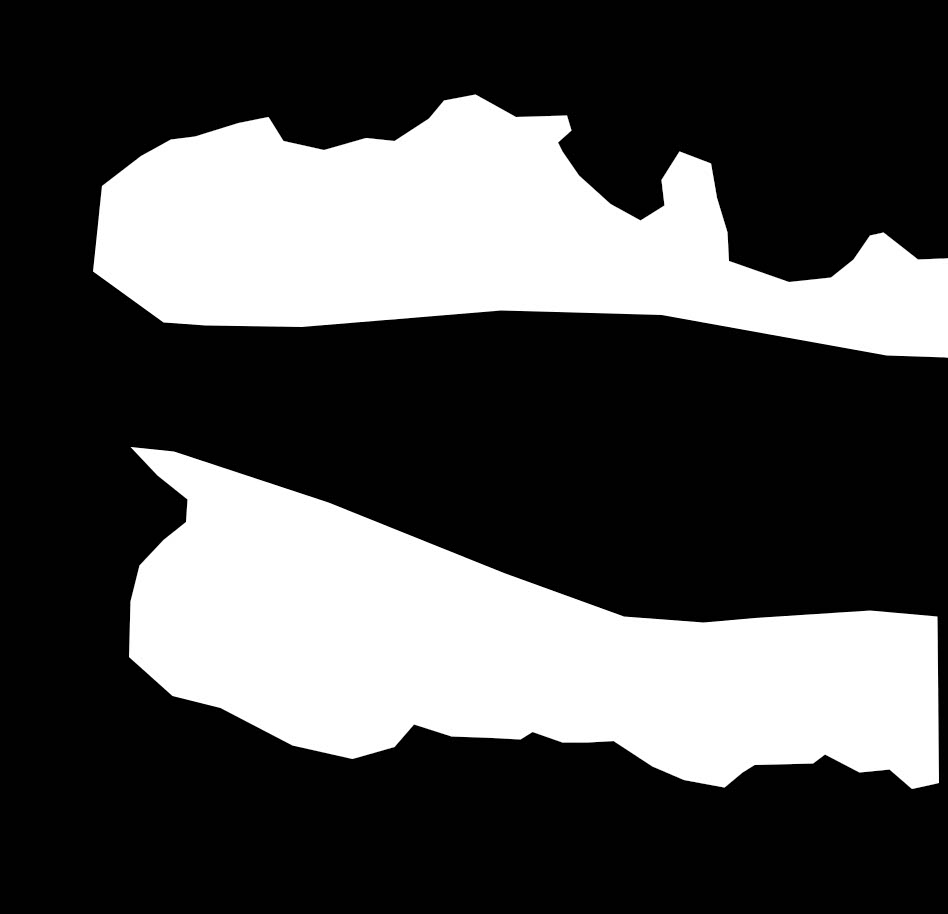}}
\hspace{1pt}
\subfloat{\includegraphics[width=0.22\textwidth]{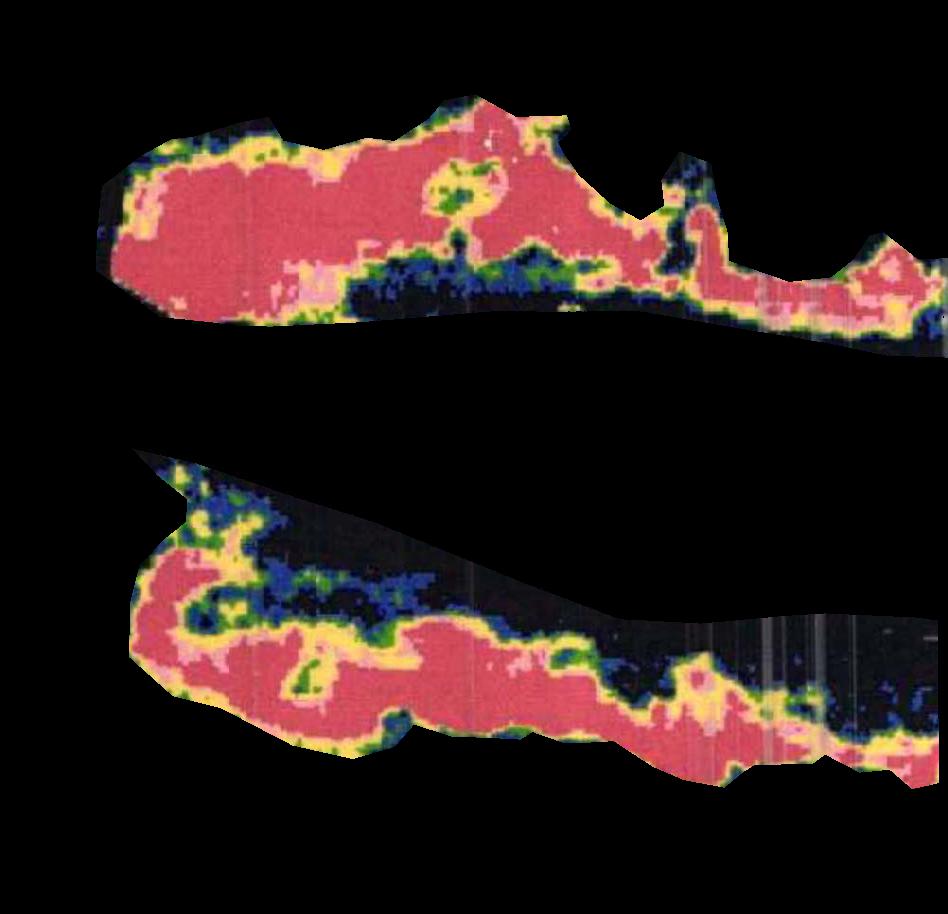}}
\hspace{1pt}
\subfloat{\includegraphics[width=0.213\textwidth]{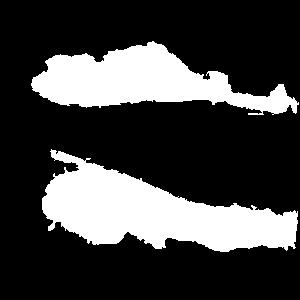}}
\\
\subfloat{\includegraphics[width=0.22\textwidth]{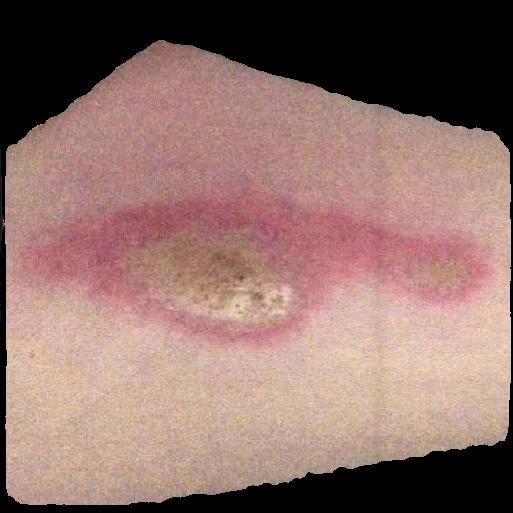}}
\hspace{1pt}
\subfloat{\includegraphics[width=0.22\textwidth]{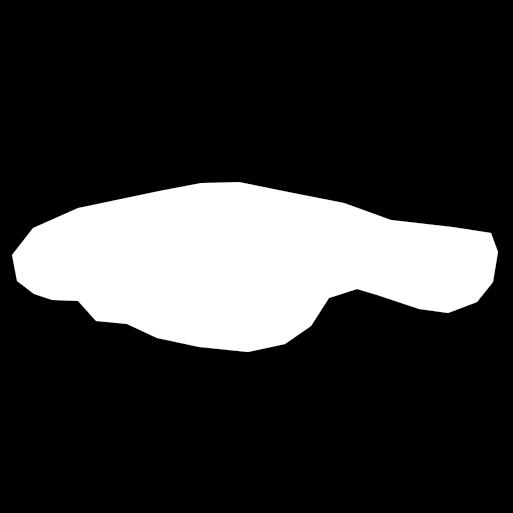}}
\hspace{1pt}
\subfloat{\includegraphics[width=0.22\textwidth]{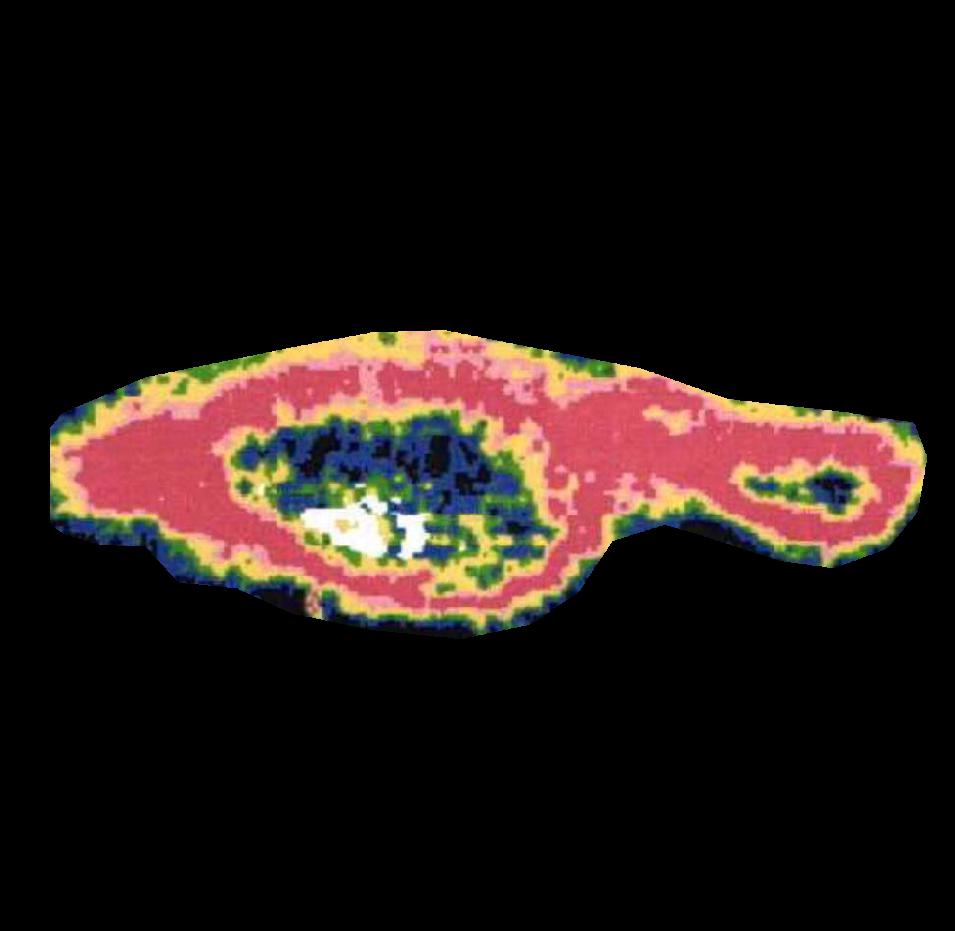}}
\hspace{1pt}
\subfloat{\includegraphics[width=0.215\textwidth]{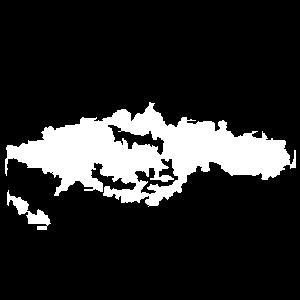}}
\\
\caption{Comparison of input images, manually segmented masks, LDI scans, and BAM maps from three severe burn patients.}
\label{fig8}
\end{center}
\end{figure}

Table \ref{table2} (top) reports four different metrics for comparing BAM segmentations with manual segmentations and LDI masks. Briefly, pixel-wise accuracy reports the ratio of correctly classified pixels to the total pixels. Pixel-wise sensitivity quantifies the ratio of correctly classified burn pixels to all the actual burn pixels representing the true positive rates. The pixel-wise specificity measures the ratio of correctly classified non-burn pixels to all the actual non-burn pixels representing the true negative rates. Finally, Jaccard Index/IOU measures the degree of overlap between ground truth segmentations (here manual segmentations or LDI segmentations) and predicted segmentations (here BAM segmentations). It is an important measure of performance as it considers both false positives and false negatives.

In addition to evaluating the BAM segmentations against the manual segmentations and the LDI scans, we examined how much improvement the BAM segmentations achieve in comparison to the Grad-CAM heatmaps. The reason for this comparison is the fact that the BAM heatmaps are generated based on the Grad-CAM heatmaps. Therefore, if the Grad-CAM heatmap fails to correctly identify the burn region in the image, BAM segmentations will also fail in generating a high-resolution heat map that highlights the burn region. In contrast, if the Grad-CAM heatmap highlights the correct region (even partially), then the BAM heatmap will be able to generate a high-resolution heatmap that highlights the burn.

Table \ref{table2} (bottom) reports the same four metrics when comparing the Grad-CAM heatmaps with the manual segmentations as well as LDI scans. For the purposes of performing this evaluation, we convert the Grad-CAM heatmaps into binary segmentations by masking them at $th=0.2$. It is evident from Table \ref{table2} that significant improvement of Jaccard Index is achieved by moving from the Grad-CAM segmentations to the BAM segmentations. Additionally, it is shown that for Grad-CAM segmentations specificity is very high while sensitivity is very low. This means that Gradm-CAM segmentations are good at partially highlighting the burn area only. However, BAM segmentations improve these partial segmentations and therefore achieve a better balance of sensitivity and specificity.

\section{Discussion}
\label{sec5}

Burn patient management is initiated with assessments that characterize the burn injury. Two critical assessments include burn severity and spatial area of burn. Identifying the severity of the injury (SPF, SPT, DPT, or FT) is important for assessing the impact the burn had on the tissue in terms of depth. The spatial area affected, or the total body surface area (TBSA\%) of the burn, helps demarcate injured versus healthy tissue and is critical in determining the resuscitative measures required for treatment. Therefore, mapping the boundary of the injury is paramount. From a medical standpoint, initial assessments that delineate superficial partial thickness (SPT) from deep partial thickness (DPT) burns are important, as it dictates the downstream, definitive treatment protocol, including transitioning a patient to specialized burn units. Spatial boundaries of the injuries also dictate resuscitation fluid administration and surgical management. 

These two initial assessments also have implications on rehabilitative activity post primary treatment, especially if the burn is over joints such as scar and range-of-motion management. Clinical assessment accuracy, depending on the expertise of the attending physician, and the presentation of the burn, can range from 60\%-80\% \cite{pape2001audit}. The clinical non-invasive gold standard for making these two assessments, including predicting the days-to-heal or healing potential, is Laser Doppler Imaging (LDI) which can have up to 97\% accuracy \cite{thatcher2016imaging}\cite{pape2001audit}. However, apart from equipment and maintenance costs, LDI also requires specialized training. For these reasons, LDI is relatively inaccessible for most physicians and burn patients. Another limitation of LDI that was highlighted during this study was the high incidence of false positive and false negative signals. This made direct interpretation of depth severity and spatial boundaries from LDI scans very inaccurate even if healing potential predictions were accurate.  

To address these challenges in burn management, we built a machine learning pipeline that uses a convolutional neural network (CNN) that is trained on images of four severity levels of burns. We introduced the boundary attention mapping  (BAM) method which uses the coarse-grained Grad-CAM visualizations as an intermediate representation for achieving finer-grained segmentations of burn regions from skin burn 2D colour images. 

The main concept behind BAM is to use the activation channels of the first convolutional layer of a deep CNN trained for burn depth classification for the purpose of saliency mapping. We first propose an iterative approach for combining the activation channels of the first convolutional layer in order to obtain a high-resolution visualization that is highly correlated to the Grad-CAM visualization. Secondly, we demonstrated that this visualization can easily be converted into a fine-grained segmentation of burn regions. Lastly, we showed the effectiveness of the BAM method through extensive qualitative results and quantitative evaluations using a skin burn image dataset and a benchmark LDI dataset.

We provide evidence that the fine-grained segmentations of burn regions from skin burn images can be used for localizing the abnormality area, which can help calculate the percentage of total body surface area (TBSA\%) that is affected by the burn. TBSA\% is an important metric for determining treatment steps and therefore finding a fast, accurate, and automatic way for measuring an injury has the potential to positively affect the clinical decision process. 

Future directions to pursue include improving the class-discriminative power of Grad-CAM visualisations since BAM integrates and depends on the steps of the Grad-CAM. In other words, BAM is not able to perform well if the Grad-CAM heatmap fails to highlight the correct “attention” region. Another possible approach would be to explore and examine the use of other types of coarse-grained class attention mapping methods instead of Grad-CAM heatmaps or even back propagation methods, like Layer-wise Relevance Propagation (LRP) \cite{ayhan2022clinical}. This can result in finding the attention mapping method best suitable for the specific application of segmenting burn regions from skin burn images. Lastly, the use of metrics other than the correlation for measuring the similarity between a coarse-grained Grad-CAM visualization and the high-resolution visualization may be examined. 

From a clinical perspective, LDI has errors (false positive and false negative signals) thus underestimating the power of the CNN-BAM system. A revision using exclusion criteria for erroneous LDI scans may give a more accurate correlation between the BAM and LDI methods of ascertaining burn depth severity and healing for any prospective study. It would also be of significant value to understand the demographics, clinical assessments, and ground-truth outcomes by following patient charts and biopsies of the patients whose LDI data was used in this study. These are being investigated in a complementary clinical study with 144 patients, and approximately 176 LDI scans, including a comparison between clinical versus LDI versus AI assessments of burn severity.

\bibliographystyle{unsrtnat}

\begin{thebibliography}{15}
\providecommand{\natexlab}[1]{#1}
\providecommand{\url}[1]{\texttt{#1}}
\expandafter\ifx\csname urlstyle\endcsname\relax
  \providecommand{\doi}[1]{doi: #1}\else
  \providecommand{\doi}{doi: \begingroup \urlstyle{rm}\Url}\fi

\bibitem[Pencle et~al.(2017)Pencle, Mowery, and Zulfiqar]{pencle2017first}
Fabio~J Pencle, Myles~L Mowery, and Hassam Zulfiqar.
\newblock First degree burn.
\newblock 2017.

\bibitem[Schaefer and Tannan(2021)]{schaefer2021thermal}
Timothy~J Schaefer and Shruti~C Tannan.
\newblock Thermal burns.
\newblock In \emph{StatPearls [Internet]}. StatPearls Publishing, 2021.

\bibitem[Cirillo et~al.(2019)Cirillo, Mirdell, Sj{\"o}berg, and
  Pham]{cirillo2019time}
Marco~Domenico Cirillo, Robin Mirdell, Folke Sj{\"o}berg, and Tuan~D Pham.
\newblock Time-independent prediction of burn depth using deep convolutional
  neural networks.
\newblock \emph{Journal of Burn Care \& Research}, 40\penalty0 (6):\penalty0
  857--863, 2019.

\bibitem[Hop et~al.(2013)Hop, Hiddingh, Stekelenburg, Kuipers, Middelkoop,
  Nieuwenhuis, Polinder, van Baar, and Group]{hop2013cost}
M~Jenda Hop, Jakob Hiddingh, Carlijn~M Stekelenburg, Hedwig~C Kuipers, Esther
  Middelkoop, Marianne~K Nieuwenhuis, Suzanne Polinder, Margriet~E van Baar,
  and LDI~Study Group.
\newblock Cost-effectiveness of laser doppler imaging in burn care in the
  netherlands.
\newblock \emph{BMC surgery}, 13:\penalty0 1--7, 2013.

\bibitem[Pape et~al.(2001)Pape, Skouras, and Byrne]{pape2001audit}
Sarah~A Pape, Costas~A Skouras, and Phillip~O Byrne.
\newblock An audit of the use of laser doppler imaging (ldi) in the assessment
  of burns of intermediate depth.
\newblock \emph{Burns}, 27\penalty0 (3):\penalty0 233--239, 2001.

\bibitem[Abubakar et~al.(2020)Abubakar, Ugail, and
  Bukar]{abubakar2020assessment}
Aliyu Abubakar, Hassan Ugail, and Ali~Maina Bukar.
\newblock Assessment of human skin burns: a deep transfer learning approach.
\newblock \emph{Journal of Medical and Biological Engineering}, 40\penalty0
  (3):\penalty0 321--333, 2020.

\bibitem[Khan et~al.(2020)Khan, Butt, Asif, Aljuaid, Adnan, Shaheen,
  et~al.]{khan2020burnt}
Fakhri~Alam Khan, Ateeq Ur~Rehman Butt, Muhammad Asif, Hanan Aljuaid, Awais
  Adnan, Sadaf Shaheen, et~al.
\newblock Burnt human skin segmentation and depth classification using deep
  convolutional neural network (dcnn).
\newblock \emph{Journal of Medical Imaging and Health Informatics}, 10\penalty0
  (10):\penalty0 2421--2429, 2020.

\bibitem[Ethier et~al.(2022)Ethier, Chan, Abdolahnejad, Morzycki, Tchango,
  Joshi, Wong, and Hong]{ethier2022using}
Olivier Ethier, Hannah~O Chan, Mahla Abdolahnejad, Alexander Morzycki,
  Arsene~Fansi Tchango, Rakesh Joshi, Joshua~N Wong, and Collin Hong.
\newblock Using computer vision and artificial intelligence to track the
  healing of severe burns.
\newblock \emph{medRxiv}, pages 2022--12, 2022.

\bibitem[Selvaraju et~al.(2017)Selvaraju, Cogswell, Das, Vedantam, Parikh, and
  Batra]{selvaraju2017grad}
Ramprasaath~R Selvaraju, Michael Cogswell, Abhishek Das, Ramakrishna Vedantam,
  Devi Parikh, and Dhruv Batra.
\newblock Grad-cam: Visual explanations from deep networks via gradient-based
  localization.
\newblock In \emph{Proceedings of the IEEE international conference on computer
  vision}, pages 618--626, 2017.

\bibitem[Tan and Le(2019)]{tan2019efficientnet}
Mingxing Tan and Quoc Le.
\newblock Efficientnet: Rethinking model scaling for convolutional neural
  networks.
\newblock In \emph{International conference on machine learning}, pages
  6105--6114. PMLR, 2019.

\bibitem[Deng et~al.(2009)Deng, Dong, Socher, Li, Li, and
  Fei-Fei]{deng2009imagenet}
Jia Deng, Wei Dong, Richard Socher, Li-Jia Li, Kai Li, and Li~Fei-Fei.
\newblock Imagenet: A large-scale hierarchical image database.
\newblock In \emph{2009 IEEE conference on computer vision and pattern
  recognition}, pages 248--255. Ieee, 2009.

\bibitem[Med(2021)]{Medtech2021moorLDLS}
moorldls-bi for burn depth assessment.
\newblock \emph{Medtech innovation briefing [MIB251]}, 2021.

\bibitem[Hoeksema et~al.(2009)Hoeksema, Van~de Sijpe, Tondu, Hamdi,
  Van~Landuyt, Blondeel, and Monstrey]{hoeksema2009accuracy}
Henk Hoeksema, Karlien Van~de Sijpe, Thiery Tondu, Moustapha Hamdi, Koenraad
  Van~Landuyt, Phillip Blondeel, and Stan Monstrey.
\newblock Accuracy of early burn depth assessment by laser doppler imaging on
  different days post burn.
\newblock \emph{Burns}, 35\penalty0 (1):\penalty0 36--45, 2009.

\bibitem[Thatcher et~al.(2016)Thatcher, Squiers, Kanick, King, Lu, Wang, Mohan,
  Sellke, and DiMaio]{thatcher2016imaging}
Jeffrey~E Thatcher, John~J Squiers, Stephen~C Kanick, Darlene~R King, Yang Lu,
  Yulin Wang, Rachit Mohan, Eric~W Sellke, and J~Michael DiMaio.
\newblock Imaging techniques for clinical burn assessment with a focus on
  multispectral imaging.
\newblock \emph{Advances in wound care}, 5\penalty0 (8):\penalty0 360--378,
  2016.

\bibitem[Ayhan et~al.(2022)Ayhan, K{\"u}mmerle, K{\"u}hlewein, Inhoffen,
  Aliyeva, Ziemssen, and Berens]{ayhan2022clinical}
Murat~Se{\c{c}}kin Ayhan, Louis~Benedikt K{\"u}mmerle, Laura K{\"u}hlewein,
  Werner Inhoffen, Gulnar Aliyeva, Focke Ziemssen, and Philipp Berens.
\newblock Clinical validation of saliency maps for understanding deep neural
  networks in ophthalmology.
\newblock \emph{Medical Image Analysis}, 77:\penalty0 102364, 2022.

\end{thebibliography}

\end{document}